\documentclass{article}
\usepackage[utf8]{inputenc}
\usepackage{mathtools}
\usepackage[a4paper,top=3cm,bottom=2cm,left=2.3cm,right=2.3cm,marginparwidth=1.75cm]{geometry}

\usepackage{amssymb}
\usepackage{natbib}
\usepackage{graphicx}
\usepackage{verbatim}
\usepackage{animate}
\usepackage{hyperref}       
\usepackage{url}            
\usepackage{booktabs}       
\usepackage{amsfonts}       
\usepackage{nicefrac}       
\usepackage{microtype}      
\usepackage{amsmath}
\usepackage{multirow}
\usepackage{amsthm}
\newtheorem{theorem}{Theorem}[section]

\newtheorem{lemma}{Lemma}[section]
\newtheorem{assumption}{Assumption}[section]
\newtheorem{definition}{Definition}[section]

\usepackage{subfigure}
\usepackage{times}
\usepackage{color}
\newcommand{\bx}{\boldsymbol{x}}
\newcommand{\bt}{\boldsymbol{\theta}}
\usepackage{algorithm}
\usepackage{algpseudocode}
\usepackage{grffile}

\begin{document}

\title{Augmented Physics-Informed Neural Networks (APINNs): A gating network-based soft domain decomposition methodology}
\author{Zheyuan Hu\thanks{Department of Computer Science, National University of Singapore, Singapore, 119077 (\href{mailto:e0792494@u.nus.edu}{e0792494@u.nus.edu},\href{mailto:kenji@nus.edu.sg}{kenji@nus.edu.sg})}
\and Ameya D. Jagtap\thanks{Division of Applied Mathematics, Brown University, Providence, RI 02912, USA (\href{mailto:ameya\_jagtap@brown.edu}{ameya\_jagtap@brown.edu}, \href{mailto:george\_karniadakis@brown.edu}{george\_karniadakis@brown.edu})}
\and George Em Karniadakis\footnotemark[2] \and  \linebreak Kenji Kawaguchi\footnotemark[1]}

\date{}
\maketitle
\begin{abstract}

In this paper, we propose the augmented physics-informed neural network (APINN), which adopts soft and trainable domain decomposition and flexible parameter sharing to further improve the extended PINN (XPINN) as well as the vanilla PINN methods.
In particular, a trainable \textit{gate network} is employed to mimic the hard decomposition of XPINN, which can be flexibly fine-tuned for discovering a potentially better partition. It weight-averages several sub-nets as the output of APINN.
APINN does not require complex interface conditions, and its sub-nets can take advantage of all training samples rather than just part of the training data in their subdomains.
Lastly, each sub-net shares part of the common parameters to capture the similar components in each decomposed function.
Furthermore, following the PINN generalization theory in \cite{hu2021extended}, we show that APINN can improve generalization by proper gate network initialization and general domain \& function decomposition.
Extensive experiments on different types of PDEs demonstrate how APINN improves the PINN and XPINN methods.
Specifically, we present examples where XPINN performs similarly to or worse than PINN, so that APINN can significantly improve both. We also show cases where XPINN is already better than PINN, so APINN can still slightly improve XPINN.
Furthermore, we visualize the optimized gating networks and their optimization trajectories, and connect them with their performance, which helps discover the possibly optimal decomposition.
Interestingly, if initialized by different decomposition, the performances of corresponding APINNs can differ drastically.
This, in turn, shows the potential to design an optimal domain decomposition for the differential equation problem under consideration.
\end{abstract}

\section{Introduction}
Deep learning has become popular in scientific computing and is widely adopted in solving forward and inverse problems involving partial differential equations (PDEs). The physics-informed neural network (PINN) \cite{raissi2019physics} is one of the seminal works in utilizing deep neural networks to approximate PDE solutions by optimizing them to satisfy the data and physical laws governed by the PDE. Furthermore, the extended PINN (XPINN) \cite{jagtap2020extended} is a follow-up work of PINN, which first proposes space-time domain decomposition to partition the domain into several subdomains, where several sub-nets are employed to approximate the solution on their subdomains, while the solution continuity between them is enforced via interface losses. Then, its output is the ensemble of all sub-nets. The theoretical analysis of when XPINNs can improve generalization over PINNs is of great interest. The recent work of Hu et al., \cite{hu2021extended} analyzes the trade-off in XPINN generalization between the simplicity of decomposed
target function in each subdomain and the overfitting effect due to less available training data in each subdomain, which counterbalance each other to determine if XPINN can improve generalization over PINN. However, sometimes the negative overfitting effect incurred by the less available training data in each subdomain dominates the positive effect of simpler partitioned target functions. Furthermore, XPINNs may also suffer from relatively large errors at the interfaces between
subdomains, which degrades the overall performance of XPINNs.

In this paper, we propose the \textit{Augmented PINN (APINN)}, which employs a gate network for soft domain partitioning to mimic the hard XPINN decomposition, which can be fine-tuned for a better decomposition. The gate network gets rid of the need for interface losses and weight-averages several sub-nets as the output of APINN, where each sub-net is able to utilize all training samples in the domain in order to prevent overfitting. Moreover, APINN adopts an efficient partial parameter sharing scheme for sub-nets, to capture the similar components in each decomposed function. 
To further understand the benefits of our APINN, we follow the theory in \cite{hu2021extended} to theoretically analyze the generalization bound of APINN, compared to those of PINN and XPINN, which justifies our intuitive understanding of the advantages of APINN. Concretely, generalization bounds for APINNs with trainable or fixed gate networks are derived, which show the advantages of soft and trainable domain and function decomposition in APINN.
We also perform extensive experiments on several PDEs that validate the effectiveness of our APINN.
Specifically, we have examples where XPINN performs similarly to or worse than PINN, so that APINN can significantly improve both. Moreover, we present cases where XPINN is already much better than PINN, but APINN can still slightly improve XPINN.
In addition to the superior performance of APINN, we also visualize the optimized gating networks and the optimization trajectories and then relate their shapes with their performances to select the potentially best decomposition. We show that if APINN is initialized by the optimal decomposition, then it can perform even better, which suggests strategies for designing the optimal domain decomposition for a given PDE problem.

\section{Related Work}
The PINN \cite{raissi2019physics} is one of the pioneering frameworks that employs deep learning techniques to solve forward and inverse problems governed by parametrized PDEs. PINN has been successfully used to solve many problems in the field of computational science since its initial publication; for more information, see \cite{raissi2018hidden, yang2019adversarial, jagtap2022deep, haghighat2021physics, jagtap2022deepKNN}.
The original idea of domain decomposition in the PINN method was proposed in \cite{jagtap2020conservative} for nonlinear conservation laws and named Conservative PINN (CPINNs). In subsequent work, the same authors proposed XPINN \cite{jagtap2020extended} for general space-time domain decomposition, where there is a sub-PINN on each sub-domain for fitting the target function on this sub-domain, while the continuity between sub-PINNs is enforced via additional interface losses (penalty terms). The Parallel PINN \cite{shukla2021parallel} is the follow-up work of CPINN and XPINN, where CPINN and XPINN are trained on multiple GPUs or CPUs simultaneously. Parareal PINN \cite{meng2020ppinn} decomposes a longer time domain into several short-time subdomains, which can be efficiently solved by a coarse-grained (CG) solver and PINN, so that Parareal PINN shows an obvious speedup over PINN in long-time integration PDEs. The main limitation of Parareal PINN is,  it cannot be applied to all types of PDEs. The hp-VPINN \cite{kharazmi2021hp} proposes a variational PINN method to decompose the domain when defining a new set of test functions, while the trial functions are still neural networks defined over the whole domain. DDM \cite{li2020deep} uses the Schwarz method for overlapping domain decomposition and training the sub-nets iteratively rather than in parallel like XPINNs and CPINNs. Also, \cite{mercier2021coarse} extends DDM through coarse space acceleration for improved convergence across a growing number of domains. \cite{li2022deep} also uses the Schwarz method, but the sub-nets are multi-Fourier feature networks instead.

The finite basis PINN (FBPINN) \cite{moseley2021finite} proposes dividing the domain into several small, overlapping sub-domains, with each of them using PINNs. Although FBPINN eliminates the need for interface conditions, our model differs from it in the following aspects: First, our domain decomposition is flexible and trainable, while FBPINN fixes the decomposition. Second, FPINN does not allow parameter sharing for the efficiency of sub-networks. Moreover, the overlapping subdomains in FBPINN can become computationally costly for multi-dimensional problems. The
penalty-free neural networks (PFNN) \cite{sheng2022pfnn} propose the idea of overlapping domain decomposition, which is different from our models for the same reasons as FBPINN.
The GatedPINN \cite{stiller2020large} proposes to adopt the idea of a mixture of experts (MoE) \cite{shazeer2017outrageously} to modify XPINNs. Although they also use a gate network to weight-average several sub-PINNs, their GatePINN is different from our APINN because the gate function in GatedPINN is randomly initialized, while that in our APINN is pretrained on an XPINN domain decomposition. Furthermore, they do not consider efficient parameter sharing for sub-nets to improve model expressiveness. 
\cite{dong2021local,DWIVEDI2021299} also proposes a similar idea of domain decomposition as in XPINNs. However, they use extreme learning machines (ELMs) to replace the neural networks in XPINNs, where only the parameters at the last layer are trained.
Based on variational principles and the deep Ritz method, D3M \cite{li2019d3m} further combines the Schwarz method for overlapping domain decomposition. Compared to our trainable domain decomposition, the domains in D3M are fixed during optimization.
To learn optimal modifications on the interfaces of different sub-domains, \cite{taghibakhshi2022learning} proposes using graph neural networks and unsupervised learning.
\cite{heinlein2021combining} presents a review on the domain decomposition method for numerical PDEs.

The first comprehensive theoretical analysis of PINNs as well as XPINNs for a prototypical nonlinear PDE, the Navier-Stokes equations, is proposed in \cite{Ryck2021ErrorAF}. The generalization abilities of PINNs and XPINNs are theoretically analyzed in \cite{hu2021extended} while the generalization and optimization capabilities of deep neural networks have been analyzed in the general field of deep learning in  \cite{kawaguchi2018generalization,kawaguchi2022robustness,kawaguchi2016deep,kawaguchi2019depth,xu2021optimization,kawaguchi2021theory,kawaguchi2022understanding}.


\section{Problem Definition and Background}
\subsection{Problem Definition}
We consider partial differential equations (PDEs) defined on the bounded domain $\Omega \subset \mathbb{R}^d$, with the following form:
\begin{equation}\label{eq:PDE}
\begin{aligned}
\mathcal{L}u^*(\bx)&=f(\bx) \ \text{in}\ \Omega, \qquad
u^*(\bx)=g(\bx) \ \text{on}\ \partial\Omega,
\end{aligned}
\end{equation}
For matrix norms, we denote the spectral norm by $\Vert \cdot\Vert_2$ and $l_{p,q}$ norms by $\Vert W \Vert_{p,q} = (\sum_j(\sum_k |W_{j,k}|^p)^{q/p})^{1/q}$.
In the following, we introduce the formulations of PINN \cite{raissi2019physics} and XPINN \cite{jagtap2020extended}.

\subsection{PINN and XPINN}
The PINN is motivated by optimizing neural networks to satisfy the data and physical laws governed by a PDE to approximate its solution. Given a set of $n_b$ boundary training points $\left\{\bx_{b,i}\right\}_{i=1}^{n_b}\subset\partial\Omega$ and $n_r$ residual training points $\left\{\bx_{r,i}\right\}_{i=1}^{n_r}\subset\Omega$, the ground truth PDE solution $u^*:\overline{\Omega}\rightarrow\mathbb{R}$ is approximated by the PINN model $u_{\bt}$, by minimizing the training loss containing a boundary loss and a residual loss:
\begin{equation}
R_S(\bt) = \frac{1}{n_b}\sum_{i=1}^{n_b} {|u_{\bt}(\bx_{b,i})-g(\bx_{b,i})|}^2 + \frac{1}{n_r}\sum_{i=1}^{n_r} {|\mathcal{L}u_{\bt}(\bx_{r,i})-f(\bx_{r,i})|}^2,
\end{equation}
where PINN learns boundary conditions in the first term, while learning the physical laws described by the PDEs in the second term. 

The XPINN extends PINN by decomposing the domain ${\Omega}$ into several subdomains where several sub-PINNs are employed. The continuity between each sub-PINNs is maintained via the interface loss function, and the output of XPINN is the ensemble of all sub-PINNs, where each of them makes predictions on their corresponding subdomains. 
Concretely, domain $\Omega$ is decomposed into $N_D$ subdomains as $\Omega = \cup_{i=1}^{N_D} \Omega_i$. The loss of XPINN contains the sum of the PINN losses of the sub-PINNs, including boundary and residual losses, plus the interface losses using points on the interfaces of different subdomains $
\partial\Omega_i\cap\partial\Omega_j$, where $i,j\in\left\{1,2,...,N_D\right\}$ such that  $
\partial\Omega_i\cap\partial\Omega_j\neq \emptyset$ to maintain the continuity between the two sub-PINNs $i$ and $j$. Specifically, XPINN loss for the $i$-th sub-PINN is
\begin{equation}
R_S^{i}(\bt^i) + \lambda_I \sum_{i,j:\partial\Omega_i\cap\partial\Omega_j\neq\emptyset} R_I(\bt^i, \boldsymbol{\theta}^j),
\end{equation}
where $\lambda_I$ is the weight controlling the strength of the interface loss, $\boldsymbol{\theta}^i$ is the parameters for subdomain $i$, and each $R_S^i(\boldsymbol{\theta})$ is the PINN loss for subdomain $i$ containing boundary and residual losses, i.e.,
\begin{equation}
    R_S^i(\boldsymbol{\theta}^i) = \frac{1}{n_{b,i}}\sum_{j=1}^{n_{b,i}} {|u_{\boldsymbol{\theta}^i}(\boldsymbol{x}^i_{b,j})-g(\boldsymbol{x}^i_{b,j})|}^2 + \frac{1}{n_{r,i}}\sum_{j=1}^{n_{r,i}} {|\mathcal{L}u_{\boldsymbol{\theta}^i}(\boldsymbol{x}^i_{r,j})-f(\boldsymbol{x}^i_{r,j})|}^2,
\end{equation}
where $n_{b,i}$ and $n_{r,i}$ are the number of boundary points and residual points in subdomain $i$ respectively, and $\boldsymbol{x}^i_{b,j}$ and $\bx^i_{r,j}$ are the $j$-th boundary and residual training points in subdomain $i$, respectively. Furthermore, $R_I(\bt^i, \bt^j)$ is the interface loss between the $i$-th and $j$-th subdomains based on interface training points $\{\bx^{ij}_{I,k}\}_{k=1}^{n_{I,ij}}\subset\partial\Omega_i\cap\partial\Omega_j$
\begin{equation}
\begin{aligned}
R_I(\boldsymbol{\theta}^i, \boldsymbol{\theta}^j) &= \frac{1}{n_{I,ij}} \sum_{k=1}^{n_{I,ij}}[ |u_{\boldsymbol{\theta}^i}(\boldsymbol{x}^{ij}_{I,k})- \{\{ 
u_{\boldsymbol{\theta}^{avg}} \}\} |^2 +\\&\ |(\mathcal{L}u_{\boldsymbol{\theta}^i}(\boldsymbol{x}^{ij}_{I,k}) - f_i(\boldsymbol{x}^{ij}_{I,k}))-(\mathcal{L}u_{\boldsymbol{\theta}^j}(\boldsymbol{x}^{ij}_{I,k}) - f_j(\boldsymbol{x}^{ij}_{I,k}))|^2 ],
\end{aligned}
\end{equation}
where $\{\{ 
u_{\boldsymbol{\theta}^{avg}} \}\}  = u_{avg} \coloneqq ({u_{\boldsymbol{\theta}^i}(\boldsymbol{x}^{ij}_{I,k})+u_{\boldsymbol{\theta}^j}(\boldsymbol{x}^{ij}_{I,k})})/{2}$,  $n_{I,ij}$ is the number of interface points between the $i$-th and $j$-th subdomains, while $\boldsymbol{x}^{ij}_{I,k}$ is the $k$-th interface points between them. The first term is the average solution continuity between the $i$-th and the $j$-th sub-nets, while the second term is the residual continuity condition on the interface given by the $i$-th and the $j$-th sub-nets. We will refer to the XPINN model introduced above as \textbf{XPINNv1}, since it is exactly the model proposed in the original work of \cite{jagtap2020extended}.

In practice, XPINNv1 may exhibit relatively larger errors near the interface, i.e., the interface losses in XPINNv1 cannot necessarily maintain the continuity between different sub-PINNs. This is because the enforcement of residual continuity conditions for PDEs involving higher-order derivatives is difficult to maintain accurately due to the obvious presence of higher-order derivatives.
Therefore, \cite{de2022error} introduces enforcing the continuity of first-order derivatives between different sub-PINNs to resolve the issue:
\begin{equation}
R_A(\bt^i, \bt^j) = \frac{1}{n_{I,ij}} \sum_{k=1}^{n_{I,ij}} \sum_{m=1}^d \left| \frac{\partial u_{\bt^i}(\bx^{ij}_{I,k})}{\partial \bx_m}-\frac{\partial u_{\bt^j}(\bx^{ij}_{I,k})}{\partial \bx_m}\right|^2,
\end{equation}
where $d$ is the problem dimension, i.e., $\bx \in \mathbb{R}^d$. With this additional term on first-order derivatives, we name the corresponding XPINN model as \textbf{XPINNv2}.

\begin{figure}
\centering
\includegraphics[scale=0.5]{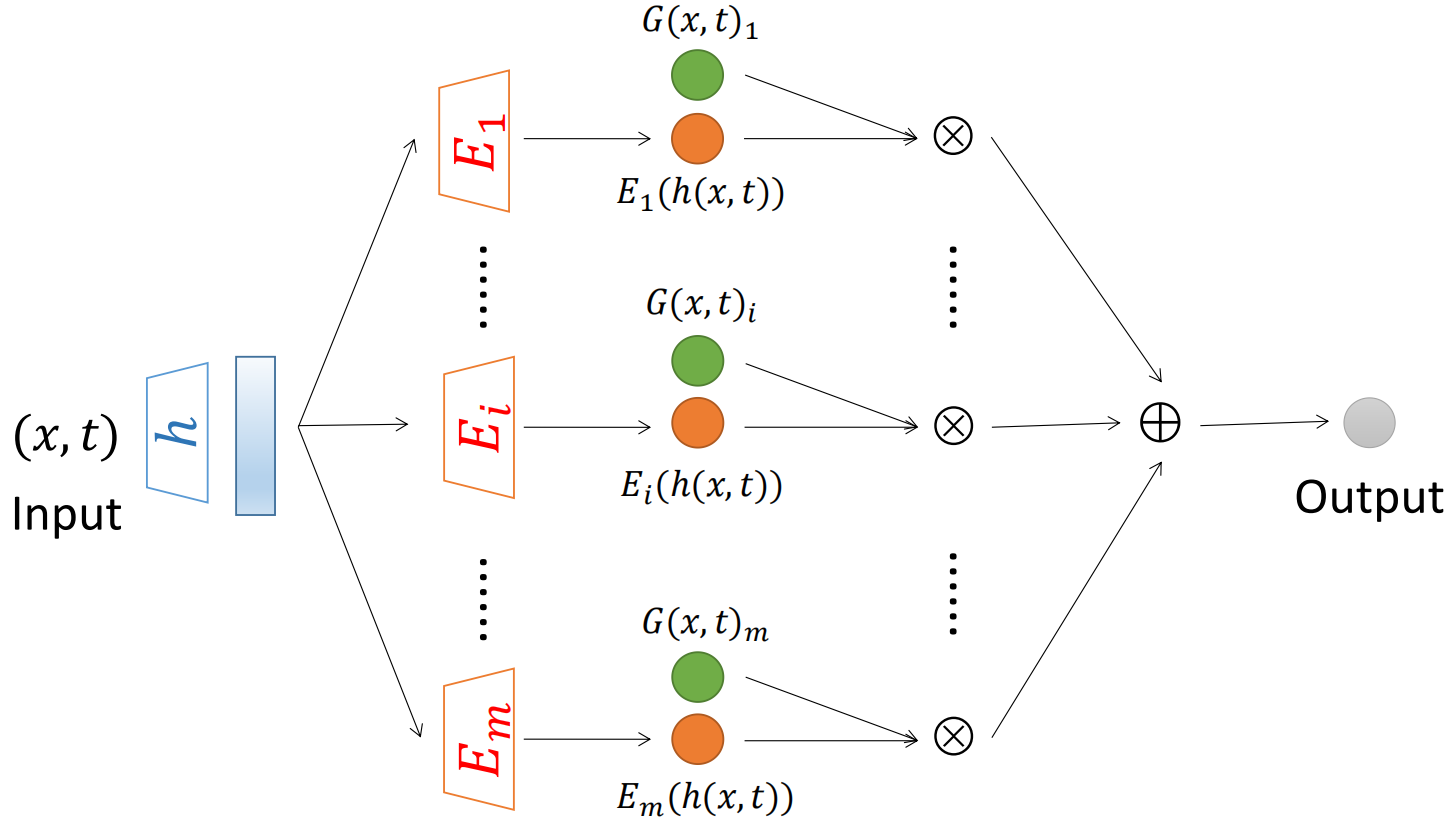}
\caption{The APINN model structure. The input $(x,t)$ is passed through the blue shared network $h$, which is then routed to $m$ distinct subnets $E_1, \cdots, E_m$ (red), yielding the corresponding $m$ outputs of subnets $E_i(h(x,t))$. Subsequently, APINN outputs the weighted average of the $m$ outputs of subnets based on the weights $G(x,t)_i$ (green), where $G$ is also a network mapping $(x,t)$ to the $m$-dimensional simplex $\Delta_m$. The weights $G(x,t)_i$ satisfies the property of partion-of-unity, i.e., $\sum_i G(x,t)_i = 1$.}
\label{fig:model}
\end{figure}

\section{Augmented PINN (APINN)}
\subsection{Parameterization of Augmented PINN}
In this section, we introduce the model parameterization of APINN, which is graphically shown in Figure \ref{fig:model}.
We consider a shared network $h: \mathbb{R}^d \rightarrow \mathbb{R}^H$ (blue), where $d$ is the input dimension and $H$ is the hidden dimension, and $m$ sub-nets $(E_i(\bx))_{i=1}^m$ (red), where each $E_i: \mathbb{R}^H \rightarrow \mathbb{R}$, and a gating network $G: \mathbb{R}^d \rightarrow \Delta^m$ (green) where $\Delta^m$ is the $m$-dimensional simplex, for weight-averaging the outputs of the $m$ sub-nets. The output of our augmented PINN (APINN) $u_{\bt}$ parameterized by $\bt$ is:
\begin{equation}
u_\theta(\bx) = \sum_{i=1}^m (G(\bx))_i E_i(h(\bx)),
\end{equation}
where $(G(\bx))_i$ is the $i$-th entry of $G(\bx)$, and $\bt$ is the collection of all parameters in $h$, $G$ and $E_i$. Both $h$ and $E_i$ are trainable in our APINN, while $G$ can be either trainable or fixed. If $G$ is trainable, we name the model APINN, otherwise we call it APINN-F.

The APINN is a universal approximator.
The detailed proof is as follows.
\begin{proof}
(The APINN is a universal approximator) Denote the function class of all neural networks as $\mathcal{NN}$, then it is universal, i.e., for all continuous functions $f \in C(\Omega)$ and $\epsilon > 0$, there exists a neural network $g \in \mathcal{NN}$, such that $\sup_{\bx \in \Omega} |f(x) - g(x)| \leq \epsilon$. In addition, we also denote the function class of gating network by $\mathcal{G}$, which collects all vector-value neural networks mapping $\mathcal{R}^d$ to $\Delta^m$.

Back to the APINN model, and denote the function class of APINN as $\mathcal{APINN}$, which is
\begin{equation}
\mathcal{APINN} = \left\{f \Big| \exists E_1, \cdots E_m, h\in \mathcal{NN}, G \in \mathcal{G}, s.t., f = \sum_{i=1}^m G(x)_i E_i(h(x))  \right\}.
\end{equation}
If we choose $E_1 = E_2 = \cdots = E_m = E$, then APINN degenerates to a vanilla multilayer network since $\sum_{i=1}^m G(x)_i = 1$, i.e., $\mathcal{NN} \subset \mathcal{APINN}$.
Therefore, since multilayer neural networks are already universal approximator and it is a subset of APINN model, APINN is a universal approximator.
\end{proof}

\begin{figure}
\centering
\includegraphics[scale=0.6]{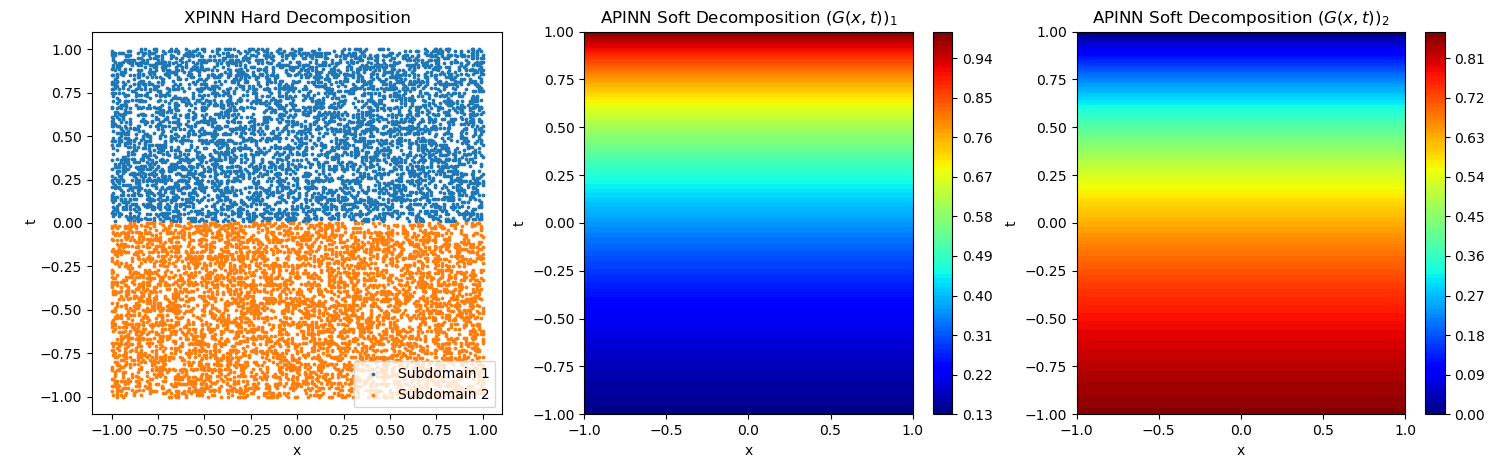}
\caption{The first example of a gating network in APINN: an upper domain (middle) and a lower domain (right).}
\label{fig:APINN1}
\end{figure}
\begin{figure}
\centering
\includegraphics[scale=0.6]{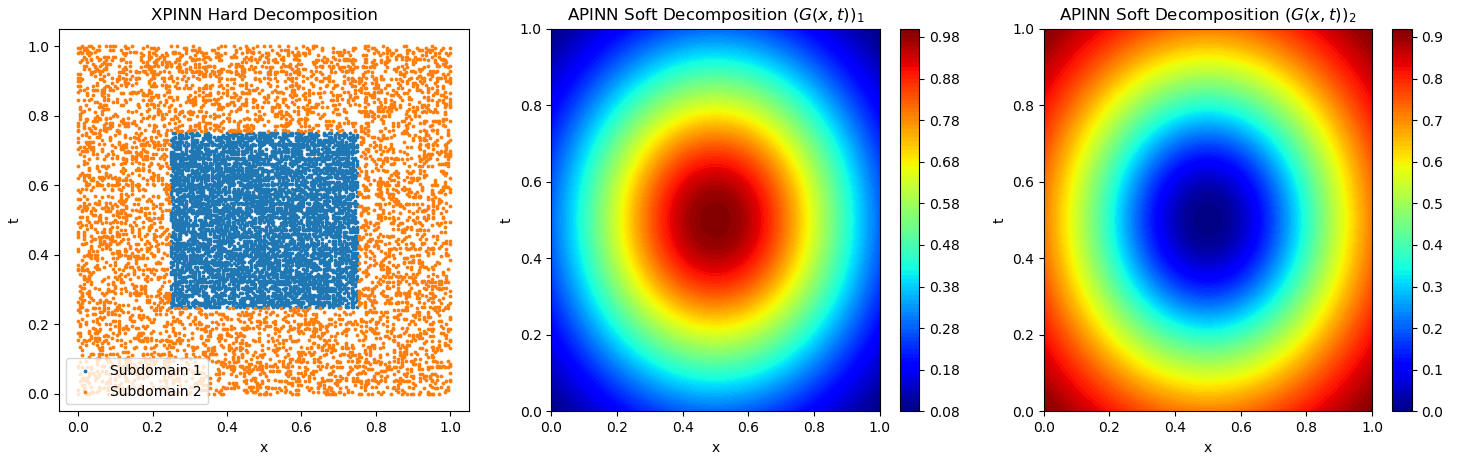}
\caption{The second example of a gating network in APINN: an inner domain (middle) and an outer domain (right).}
\label{fig:APINN2}
\end{figure}

In APINN, $G$ is pre-trained to mimic the hard and discrete decomposition of XPINN, which will be discussed in the next subsection.
If $G$ is trainable, then our model can fine-tune the pre-trained domain decomposition to further discover a better decomposition through optimization. If not, then APINN is exactly the soft version of XPINN with the corresponding hard decomposition.
APINN is better than PINN thanks to the adaptive domain decomposition and parameter efficiency.

\subsection{Explanation of the Gating Network}
In this section, we will show how the gating network $G$ can be trained to mimic XPINNs for soft domain decomposition.
Specifically, in Figure \ref{fig:APINN1} left, XPINN decomposes the entire domain $(x,t) \in \Omega = [-1, 1] \times [-1,1]$, into two subdomains: the upper one $\Omega_1 = [-1, 1] \times [0,1]$, and the lower one $\Omega_2 = [-1, 1] \times [-1,0)$, which is based on the interface $t = 0$.
The soft domain decomposition in APINN is shown in Figure \ref{fig:APINN1} (middle and right), which are the pretrained gating networks for the two sub-nets corresponding to the upper and bottom subdomains. Here, $(G(x,t))_1$ is pretrained on $\exp(t-1)$ and $(G(x,t))_2$ on $1 - \exp(t-1)$. Intuitively, the first sub-PINN focuses on where $t$ is larger, corresponding to the upper part, while the second sub-PINN focuses on where $t$ is smaller, corresponding to the bottom part.

Another example is to decompose the domain into an inner part and an outer part, as shown in Figure \ref{fig:APINN2}. 
In particular, we decompose the entire domain $(x,t) \in \Omega = [0, 1] \times [0,1]$, into two subdomains: the inner one $\Omega_1 = [0.25, 0.75] \times [0.25,0.75]$, and the outer one $\Omega_2 = \Omega \setminus \Omega_1$.
The soft domain decomposition is generated by the gating functions $(G(x,t))_1$ pretrained on $\exp(-5(x - 0.5)^2 - 5(t-0.5)^2)$ and $(G(x,t))_2$ pretrained on $1 - \exp(-5(x - 0.5)^2 - 5(t-0.5)^2)$, such that the first sub-net concentrates in the inner part near $(x,t)=(0.5,0.5)$, while the second sub-bet focuses on the rest of the domain.

The gating network can also be adapted for complex domains like the L-shape domain or even high-dimensional domains by properly choosing the corresponding gating function.

\subsection{Difference in the position of $h$}
We have three options for building the model of APINN. First, the simplest idea is that if we omit the parameter sharing in our APINN, then the model becomes:
\begin{equation}\label{eq:model1}
u_\theta(\bx) = \sum_{i=1}^m (G(\bx))_i E_i(\bx).
\end{equation}
The proposed model in this paper is
\begin{equation}\label{eq:model2}
u_\theta(\bx) = \sum_{i=1}^m (G(\bx))_i E_i(h(\bx)).
\end{equation}
Another method for parameter sharing is to place $h $ outside the weighted average of several sub-nets.
\begin{equation}\label{eq:model3}
u_\theta(\bx) = h\left(\sum_{i=1}^m (G(\bx))_i E_i(\bx)\right).
\end{equation}
Compared to the first model, our new model given in equation (\ref{eq:model2}) adopts parameter sharing for each sub-PINN to improve parameter efficiency. Equation (\ref{eq:model2}) generalizes equation (\ref{eq:model1}) by using the same $E_i$ networks and selecting the shared network as identity mapping. Intuitively, the functions learned by each sub-PINNs should have some kind of similarity, since they are parts of the same target function. The prior of network sharing in our model explicitly utilizes intuition and is therefore more parameter efficient.

Compared to the model given in equation (\ref{eq:model3}), our model is more interpretable. In particular, our model in equation \ref{eq:model2} is a weighted average of $m$ sub-PINNs $E_i \circ h$, so that we can visualize each $E_i \circ h$ to observe what functions they are learning. However, for equation (\ref{eq:model3}), there is no clear function decomposition due to the $h$ being outside, so that visualization of each function component learned is not possible.

\section{Theoretical Analysis}

\subsection{Preliminaries}
To facilitate the statement of our main generalization bound, we first define several quantities related to the network parameters. For a network $u_{\bt}(x)=W_L \sigma (W_{L-1} \sigma(\cdots \sigma(W_1x)\cdots)$, we denote $M(l) = \lceil \Vert W_l \Vert_2 \rceil$ and $N(l) = \lceil \frac{\Vert W_l - A_l \Vert_{2,1}}{\Vert W_l \Vert_2} \rceil$ for fixed reference matrices $A_l$, where $A_l$ can vary for different networks. We denote its complexity as follows
\begin{equation}
R_i(u_{\bt}) = \left(\prod_{l=1}^L M(l)\right)^{i+1} \left(\sum_{l=1}^L N(l)^{2/3}\right)^{3/2}, \quad i \in \{0,1,2\},
\end{equation}
where $i$ signifies the order of derivative, i.e., $R_i$ denotes the complexity of the $i$-th derivative of the network.
We further denote the corresponding $M(l)$, $N(l)$, and $R_i$ quantities of the sub-PINN $E_j\circ h$ as $M_j(j)$, $N_j(l)$ and $R_i(E_j \circ G)$. We also denote those of the gate network $G$ as $M_G(l)$, $N_G(l)$ and $R_i(G)$.



The train loss and test loss of a model $u_{\bt}(\bx)$ are the same as that of PINN, i.e.,
\begin{equation}
\begin{aligned}
R_S(\bt) &= R_{S\cap\partial\Omega}(\bt) +R_{S\cap\Omega}(\bt) \\
&=\frac{1}{n_b}\sum_{i=1}^{n_b} {|u_{\boldsymbol{\theta}}(\boldsymbol{x}_{b,i})-g(\boldsymbol{x}_{b,i})|}^2 + \frac{1}{n_r}\sum_{i=1}^{n_r} {|\mathcal{L}u_{\boldsymbol{\theta}}(\boldsymbol{x}_{r,i})-f(\boldsymbol{x}_{r,i})|}^2.\\
R_{{D}}(\boldsymbol{\theta})&= R_{D\cap\partial\Omega}(\bt) +R_{D\cap\Omega}(\bt) \\&=\mathbb{E}_{\text{Unif}(\partial\Omega)} {|u_{\boldsymbol{\theta}}(\boldsymbol{x})-g(\boldsymbol{x})|}^2 + \mathbb{E}_{\text{Unif}(\Omega)} {|\mathcal{L}u_{\boldsymbol{\theta}}(\boldsymbol{x})-f(\boldsymbol{x})|}^2.
\end{aligned}
\end{equation}
Since the following assumption holds for a vast variety of PDEs, we can bound the test $L_2$ error by the test boundary and residual losses:
\begin{assumption}\label{assumption:L2}
Assume that the PDE satisfies the following norm constraint:
\begin{equation}
\begin{aligned}
C_1 \Vert u \Vert_{L_2(\Omega)} \leq \Vert \mathcal{L}u \Vert_{L_2(\Omega)} + \Vert u \Vert_{L_2(\partial\Omega)}, \qquad \forall u \in \mathcal{NN}_{L}, \forall L,
\end{aligned}
\end{equation}
where the positive constant $C_1$ does not depend on $u$ but on the domain and the coefficients of the operators $\mathcal{L}$, and the function class $\mathcal{NN}_{L}$ contains all $L$-layer neural networks.
\end{assumption}
The following assumption is widely adopted in related works \cite{Luo2020TwoLayerNN,hu2021extended}.
\begin{assumption}
\label{assumption:bounded}
(Symmetry and boundedness of $\mathcal{L}$). Throughout the analysis in this paper, we assume the differential operator $\mathcal{L}$ in the PDE satisfies the following conditions.
The operator $\mathcal{L}$ is a linear second-order differential operator in a non-divergence form, i.e.,
$(\mathcal{L}u^*)(\bx) = \sum_{\alpha=1,\beta=1}^d \boldsymbol{A}_{\alpha \beta} (\bx)u^*_{x_\alpha x_\beta} (\bx)+ \sum_{\alpha=1}^d \boldsymbol{b}_\alpha (\bx)u^*_{x_\alpha} (\bx)+ c(\bx) u^*(\bx)$,
where all $\boldsymbol{A}_{\alpha \beta},\boldsymbol{b}_\alpha, c:\Omega \rightarrow \mathbb{R}$ are given coefficient functions and $u^*_{x_\alpha}$ are the first-order partial derivatives of the function $u^*$ with respect to its $\alpha$-th argument (the variable $x_\alpha$) and $u^*_{x_\alpha x_\beta}$ are the second-order partial derivatives of the function $u^*$ with respect to its $\alpha$-th and $\beta$-th arguments (the variables $x_\alpha$ and $x_\beta$).
Furthermore, there exists constant $K>0$ such that for all $\bx\in\Omega=[-1,1]^d$, and $\alpha,\beta\in[d]$, we have $A_{\alpha\beta}=A_{\beta\alpha}$ and
$A_{\alpha\beta}(\boldsymbol{x}),  b_\alpha(\boldsymbol{x}),  c(\boldsymbol{x}) $ are all $K$-Lipschitz, and their absolute values are not larger than $K$.
\end{assumption}

\subsection{A Tradeoff in XPINN Generalization}
In this subsection, we review the tradeoff in XPINN generalization, introduced in \cite{hu2021extended}. There are two factors that counterbalance each other to affect XPINN generalization, namely the simplicity of the decomposed target function within each subdomain thanks to the domain decomposition, and the complexity and negative overfitting effect due to the lack of available training data. When the former effect is more obvious, XPINN outperforms PINN. Otherwise, PINN outperforms XPINN. When the two factors strike a balance, XPINN and PINN perform similarly.

\subsection{APINN with Non-Trainable Gate Network}
In this section, we state the generalization bound for APINN with a non-trainable gate network.
Since the gate network is fixed, the only complexity comes from the sub-PINNs. The following theorem holds for any gate function $G$.
\begin{theorem}\label{thm:1}
Assume that \ref{assumption:bounded} holds for any $\delta\in(0,1)$, with a probability of at least $1-\delta$ over the choice of random samples $S=\left\{\boldsymbol{x}_{i}\right\}_{i=1}^{n_b+n_r} \subset \overline{\Omega}$ with $n_b$ boundary points and $n_r$ residual points, we have the following generalization bound for an APINN model $u_{\bt_S}(\bx) = \sum_{i=1}^m (G(\bx))_i E_i(h(\bx))$:
\begin{equation}
\begin{aligned}
R_{D \cap \partial \Omega}(\bt_S) &\leq R_{S\cap \partial \Omega}(\bt_S) + \Tilde{O}\left(\frac{\sum_{j=1}^m\max_{\bx \in \partial \Omega}\Vert G(\bx)_j \Vert_{\infty}R_0(E_j \circ h)}{n_{b}^{1/2}}+ \sqrt{\frac{\log(4/\delta(E))}{n_{b}}} \right).\\
R_{D \cap \Omega}(\bt_S) &\leq R_{S\cap \Omega}(\bt_S) + \Tilde{O}\left(\frac{\sum_{i=0}^2\sum_{j=1}^m\max_{\bx \in \partial \Omega}\left\| \text{vec}\left( \frac{\partial^i G(\bx)_j}{\partial \bx^i} \right) \right\|_{\infty}R_{2-i}(E_j \circ h)}{n_{r}^{1/2}}+ \sqrt{\frac{\log(4/\delta(E))}{n_{r}}} \right), 
\end{aligned}
\end{equation}
where
$
\delta(E) = \frac{\delta}{{\prod_{l=1}^L\prod_{j \in \{1,\cdots,m\}}  M_j(l)(M_j(l)+1)N_j(l)(N_j(l)+1)}}.
$
\end{theorem}
Intuition: The first term is the train loss, and the third is the probability term, in which we divide the probability $\delta$ into $\delta(E)$ for a union bound over all parameters in $E_j \circ h$.The second term is the Rademacher complexity of the model.
For the boundary loss, the network $u_{\bt_S}(\bx) = \sum_{j=1}^m (G(\bx))_j E_j(h(\bx))$ is not differentiated. So, each $E_j(h(\bx))$ contributes $R_0(E_j \circ h)$, and $(G(\bx))_j$ contributes $\max_{\bx \in \partial \Omega}\Vert G(\bx)_j \Vert_{\infty}$ since it is fixed and is $\max_{\bx \in \partial \Omega}\Vert G(\bx)_j \Vert_{\infty}$ Lipschitz.
For the residual loss, the case of the second term is similar. Note that the second-order derivative of APINN is
\begin{equation}
\frac{\partial^2 u_{\bt_S}(\bx)}{\partial \bx^2} =\sum_{i=0}^2 \sum_{j=1}^m \frac{\partial^i (G(\bx))_j}{\partial \bx^i} \frac{\partial^{2-i}E_j(h(\bx))}{\partial \bx^{2-i}}.
\end{equation}
Consequently, each $\frac{\partial^{2-i}E_j(h(\bx))}{\partial \bx^{2-i}}$ contributes $R_{2-i}(E_j \circ h)$, while each $\frac{\partial^i (G(\bx))_j}{\partial \bx^i}$ contributes $\max_{\bx \in \partial \Omega}\left\| \text{vec}\left( \frac{\partial^i G(\bx)_j}{\partial \bx^i} \right) \right\|_{\infty}$ since it is fixed.

\subsection{Explain the Effectiveness of APINN via Theorem \ref{thm:1}}
In this section, we explain the effectiveness of APINNs using Theorem \ref{thm:1}, which shows that the benefit of APINN comes from (1) soft domain decomposition, (2) getting rid of interface losses, (3) general target function decomposition, and (4) the fact that each sub-PINN of APINN is provided with all the training data, which prevents overfitting.

For the boundary loss of APINN, we can apply Theorem \ref{thm:1} to each of the APINN's soft subdomains. Specifically, for the $k$-th sub-net in the $k$-th soft subdomain of APINN, i.e., the $\Omega_k, k\in\left\{1,2,...,m\right\}$, the bound is
\begin{equation}
R_{D \cap \Omega_k}(\bt_S) \leq R_{S \cap \Omega_k}(\bt_S)+ \Tilde{O}\left(\frac{\sum_{j=1}^m\max_{\bx \in \partial \Omega_k}\Vert G(\bx)_j \Vert_{\infty}R_0(E_j \circ h)}{n_{b,k}^{1/2}}+ \sqrt{\frac{\log(4/\delta(E))}{n_{b,k}}} \right),
\end{equation}
where $n_{b,k}$ is the number of training boundary points in the $k$-th subdomain. 

If the gate net is mimicking the hard decomposition of XPINN, then we assume that the $k$-th sub-PINN $E_k$ focuses on $\Omega_k$, in particular $\Vert G(\bx)_j \Vert_{\infty} \leq \overline{c}$ for $j \neq k$, where $\overline{c}$ approaches zero. Note that Theorem \ref{thm:1} does not depend on any requirement on the quantity $\overline{c}$, and we are making such assumption for illustration. Then, the bound reduces to
\begin{equation}
\begin{aligned}
R_{D \cap \Omega_k}(\bt_S) &\leq R_{S \cap \Omega_k}(\bt_S)+ \Tilde{O}\left(\frac{\Vert G(\bx)_k \Vert_{\infty}R_0(E_k \circ h) + \overline{c}\sum_{j \neq k}R_0(E_j \circ h)}{n_{b,k}^{1/2}}+ \sqrt{\frac{\log(4/\delta(E))}{n_{b,k}}} \right)\\
&\approx R_{S \cap \Omega_k}(\bt_S)+ \Tilde{O}\left(\frac{R_0(E_k \circ h)}{n_{b,k}^{1/2}}+ \sqrt{\frac{\log(4/\delta(E))}{n_{b,k}}} \right),
\end{aligned}
\end{equation}
which is exactly the bound of XPINN if the domain decomposition is hard.

Therefore, APINN has the benefit of XPINN, i.e., it can decompose the target function into several simpler parts in some sub-domains.
Furthermore, since APINN does not require the complex interface losses, its train loss $R_S(\bt_S)$ is usually smaller than that of XPINN, and it is free from errors near the interface.

In addition to soft domain decomposition, even if the output of $G$ does not concentrate on certain sub-domains, i.e., does not mimic XPINN, APINN still enjoys the benefit of general function decomposition, and each sub-PINN of APINN is provided with all training data, which prevents overfitting.
Concretely, for boundary loss of APINN, the complexity term of the model is $$\frac{\sum_{j=1}^m\max_{\bx \in \partial \Omega}\Vert G(\bx)_j \Vert_{\infty}R_0(E_j \circ h)}{n_b^{1/2}},$$ which is a weighted average of the complexity of all sub-PINNs. 
Note that, similar to PINN, if we view APINN on the entire domain, then all sub-PINNs are able to take advantage of all training samples, thus preventing overfitting.
Hopefully, the weighted sum of each part is simpler than the whole. To be more specific, if we train a PINN, $u_{\bt}$, the complexity term will be $R_0(u_{\bt})$.
If APINN is able to decompose the target function into several simpler parts such that their complexity weighted sum is smaller than the complexity of PINN, then APINN can outperform PINN.

\subsection{APINN with Trainable Gate Network}
In this section, we state the generalization bound for APINN with a trainable gate network. In this case, both the gate network and the $m$ sub-PINNs contribute to the complexity of the APINN model, influencing generalization at the same time. 
\begin{theorem}\label{thm:2}
Let Assumption \ref{assumption:bounded} holds, for any $\delta\in(0,1)$, with probability at least $1-\delta$ over the choice of random samples $S=\left\{\boldsymbol{x}_{i}\right\}_{i=1}^{n_b+n_r} \subset \overline{\Omega}$ with $n_b$ boundary points and $n_r$ residual points, we have the following generalization bound for an APINN model $u_{\bt_S}(\bx) = \sum_{i=1}^m (G(\bx))_i E_i(h(\bx))$:
\begin{equation}
\begin{aligned}
R_{D \cap \partial \Omega}(\bt_S) &\leq R_{S\cap \partial \Omega}(\bt_S) + \Tilde{O}\left(\frac{R_0(G)+\sum_{j=1}^m R_0(E_j \circ h)}{n_{b}^{1/4}}+ \sqrt{\frac{\log(4/\delta(G,E))}{n_{b}}} \right).\\
R_{D \cap \Omega}(\bt_S) &\leq R_{S\cap \Omega}(\bt_S) + \Tilde{O}\left(\frac{\sum_{i=0}^2\left(R_i(G)+\sum_{j=1}^mR_{2-i}(E_j \circ h)\right)}{n_{r}^{1/4}}+ \sqrt{\frac{\log(4/\delta(G,E))}{n_{r}}} \right),
\end{aligned}
\end{equation}
where
$
\delta(G,E) = \frac{\delta}{{\prod_{l=1}^L\prod_{j \in \{1,\cdots,m,G\}}  M_j(l)(M_j(l)+1)N_j(l)(N_j(l)+1)}}.
$
\end{theorem}
{Intuition}: It is somehow similar to that of Theorem \ref{thm:1}. Here, we treat the APINN model as a whole. Now, $G(\bx)$ will contribute its complexity, $R_i(G)$, rather than its infinity norm, since it is trainable rather than fixed.
\begin{figure}
\centering
\includegraphics[scale=0.5]{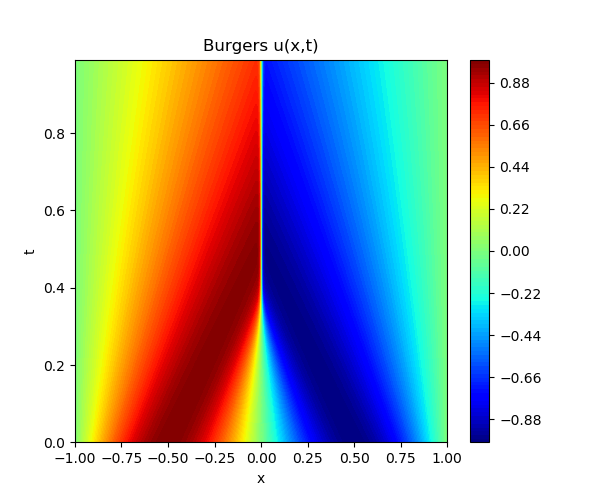}
\includegraphics[scale=0.5]{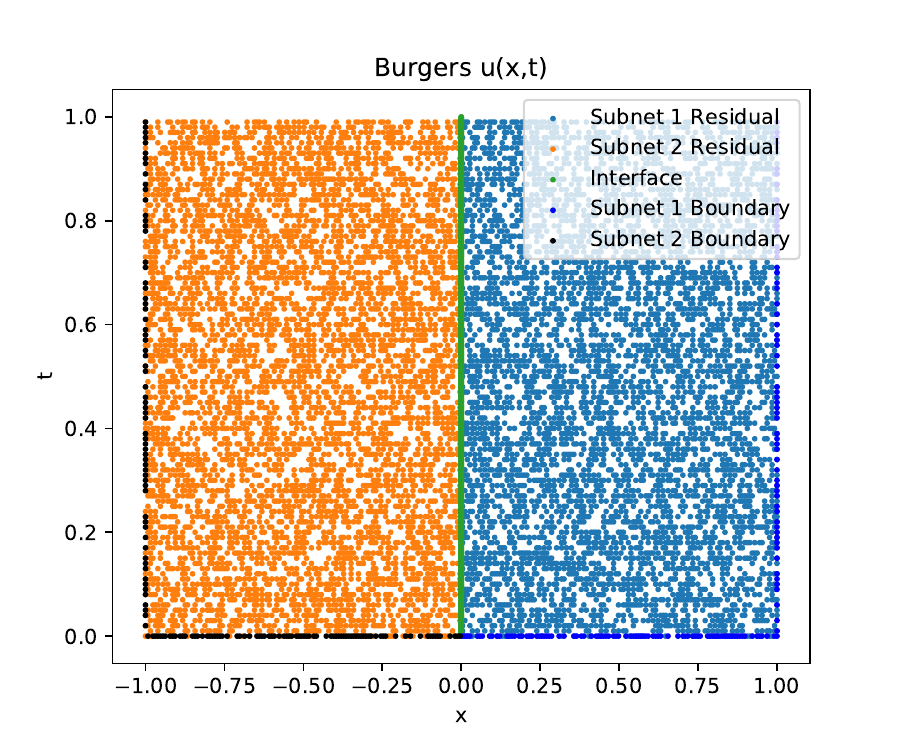}
\caption{The Burgers equation. Left: ground truth solution. Right: training points of XPINN.}
\label{fig:burgers1}
\end{figure}
\begin{figure}
\centering
\includegraphics[scale=0.8]{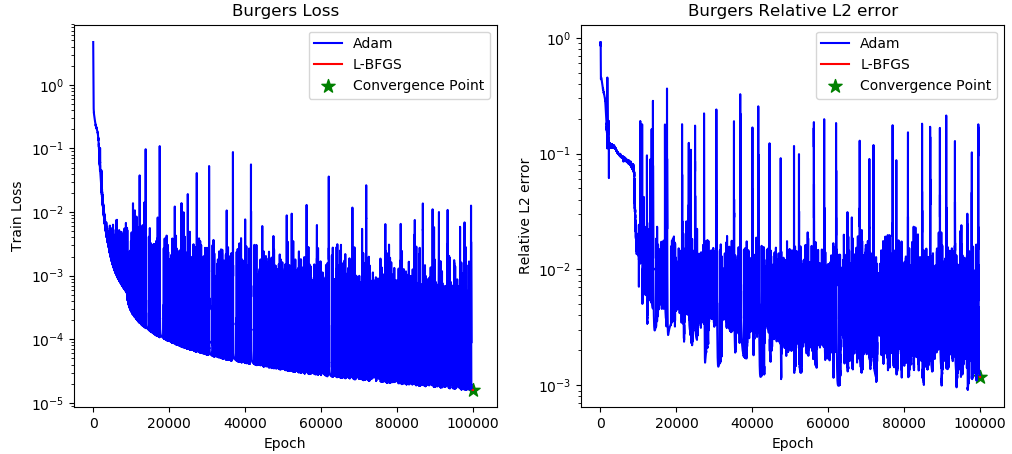}
\caption{The Burgers equation. Train loss and relative $L_2$ error. Blue: Adam optimization. Red: L-BFGS finetuning. Green: final convergence point. In this case, Adam can already train the model to convergence, so additional L-BFGS converges fast due to its stopping criterion.} 
\end{figure}
\subsection{Explain the Effectiveness of The APINN via Theorem \ref{thm:2}}
By Theorem \ref{thm:2}, besides the benefits explained by Theorem \ref{thm:1}, a good initialization of soft decomposition inspired by XPINN helps generalization. If this is the case, the trained gate network's parameters will not deviate significantly from their initialization. Consequently, $N_j(l)$ quantities for all $j \in \{1,\cdots,m,G\}$ and $l\in\{1,\cdots,L\}$ will be smaller, and thus $R_i(G)$ will be smaller, decreasing the right hand side of the bound stated in Theorem \ref{thm:2}, which means good generalization.

\section{Computational Experiments}
\subsection{The Burgers Equation}
The one-dimensional viscous Burgers equation is given by
\begin{equation}
\begin{aligned}
& u_t + uu_x - \frac{0.01}{\pi}u_{xx} = 0, x \in [-1,1], t\in [0,1].\\
& u(0,x) = -\sin(\pi x).\\
& u(t,-1) = u(t,1) = 0.
\end{aligned}
\end{equation}
The difficulty of the Burgers equation is in the steep region near $x = 0$ where the solution changes rapidly, which is hard to capture by PINNs. The ground truth solution is visualized in Figure \ref{fig:burgers1} left. In this case, XPINN performs badly near the interface. Thus, APINN improves XPINN, especially in the accuracy near the interface, both by getting rid of the interface losses and by improving the parameter efficiency.

\subsubsection{PINN and Hard XPINN}
For the PINN, we use a 10-layer tanh network of 20-width with 3441 neurons, and provide 300 boundary points and 20000 residual points. We use 20 as the weight on the boundary and 1 as the weight for the residual. We train PINN by the Adam optimizer with 8e-4 learning rate for 100k epochs.
XPINNv1 decomposes the domain based on whether $x > 0$.
The weights for boundary loss, residual loss, interface boundary loss, and interface residual loss are 20, 1, 20, 1, respectively. 
XPINNv2 shares the same decomposition as XPINNv1, but its weights for boundary loss, residual loss, interface boundary loss, and interface first-order derivative continuity loss are 20, 1, 20, 1, respectively. 
The sub-nets are 6-layer tanh networks of 20-width with 3522 neurons in total, and we provide 150 boundary points and 10000 residual points for all sub-nets in XPINN. The number of interface points is 1000. 
The training points of XPINNs are visualized in Figure \ref{fig:burgers1} right.
We train XPINNs by the Adam optimizer with 8e-4 learning rate for 100k epochs.
Both models are finetuned by the L-BFGS optimizer until convergence after Adam optimization.
\begin{table}[]
\centering
\caption{Results for the Burgers' equation.}
\label{tab:burgers}
\begin{tabular}{|c|c|c|c|c|}
\hline
Model & PINN & XPINNv1 & XPINNv2 & -\\ \hline
Rel. $L_2$ & 1.620E-3$\pm$7.632E-4 & 1.490E-1$\pm$6.781E-3 & 1.304E-1$\pm$7.256E-3 & -\\ \hline
Model & APINN-X-F & APINN-X & APINN-M-F & APINN-M \\ \hline
Rel. $L_2$ & 1.293E-3$\pm$4.629E-4 & \textbf{9.109E-4$\pm$3.689E-4} & 1.375E-3$\pm$6.732E-4 & 1.137E-3$\pm$7.675E-4 \\ \hline
\end{tabular}
\end{table}

\begin{figure}
\centering
\includegraphics[scale=0.82]{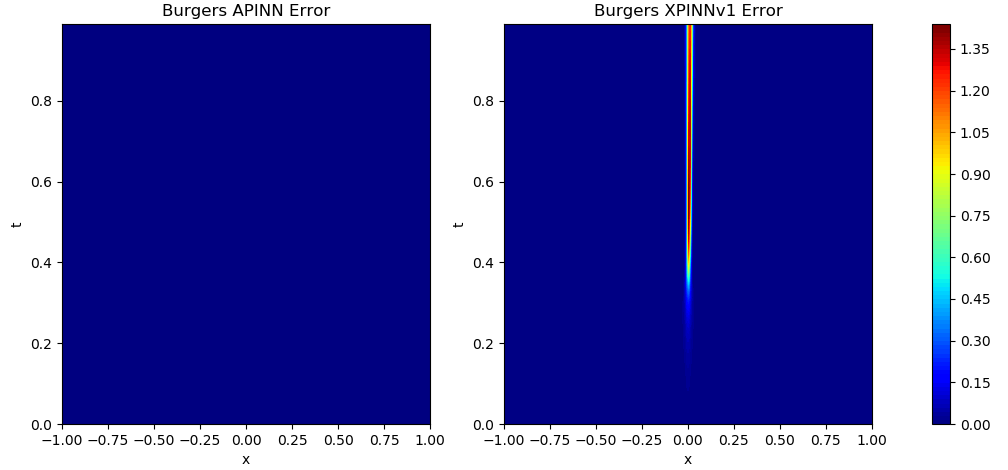}
\caption{{The Burgers equation. Left: error plot of APINN. Right: error plot of XPINNv1. Note that APINN and XPINNv1 share the same colorbar. Compared to the error of XPINNv1, that of APINN is negligible.}}
\label{fig:burgers2}
\end{figure}

\begin{figure}
\centering
\includegraphics[scale=1]{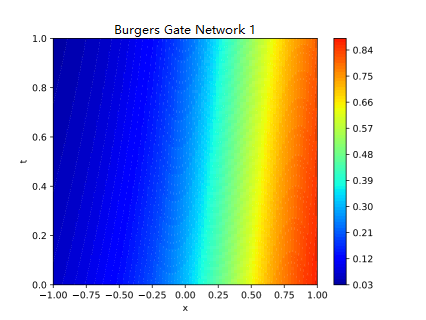}
\includegraphics[scale=1]{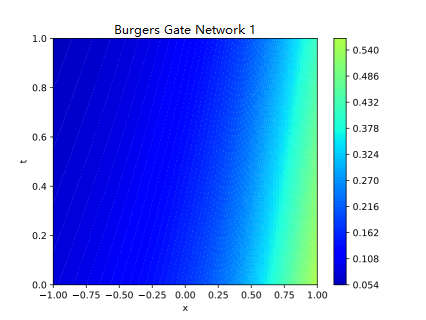}
\includegraphics[scale=1]{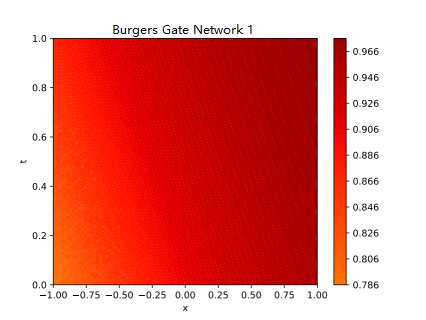}
\includegraphics[scale=1]{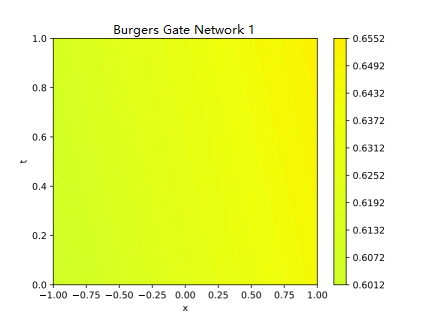}
\caption{The Burgers equation: APINN gate nets $G_1$ after convergence at the last epoch. 
That for the second subnet $G_2$ can be easily computed using the property of partition-of-unity $G_1 + G_2 = 1$.
First row: those of APINN-X with two different random seeds. Relative $L_2$ errors = 7.541E-4, 8.034E-4.
Second row: those of APINN-M with two different random seeds. Relative $L_2$ errors = 6.936E-4, 8.284E-4.}
\label{fig:burgers_gate}
\end{figure}

\begin{figure}
\centering
\includegraphics[scale=0.35]{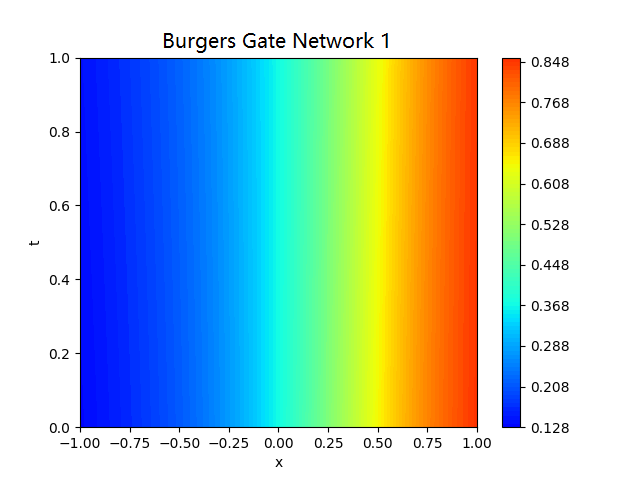}
\includegraphics[scale=0.35]{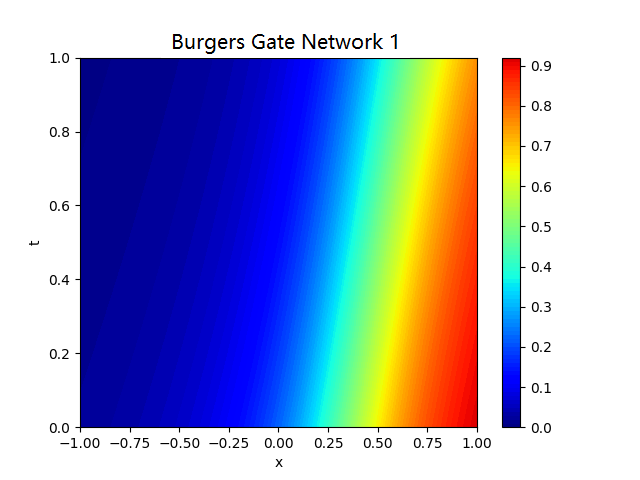}
\includegraphics[scale=0.35]{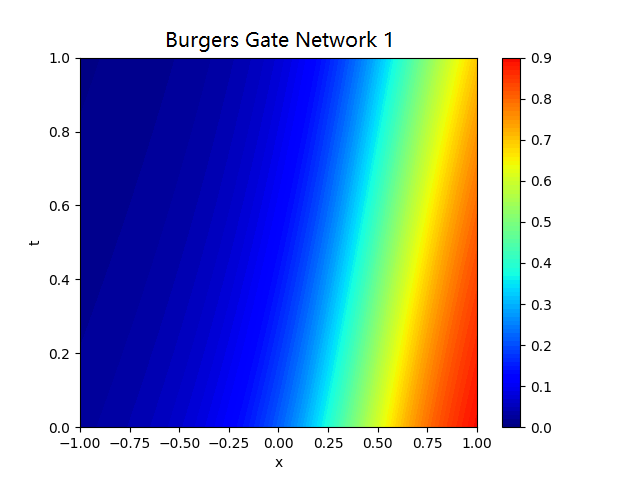}
\includegraphics[scale=0.35]{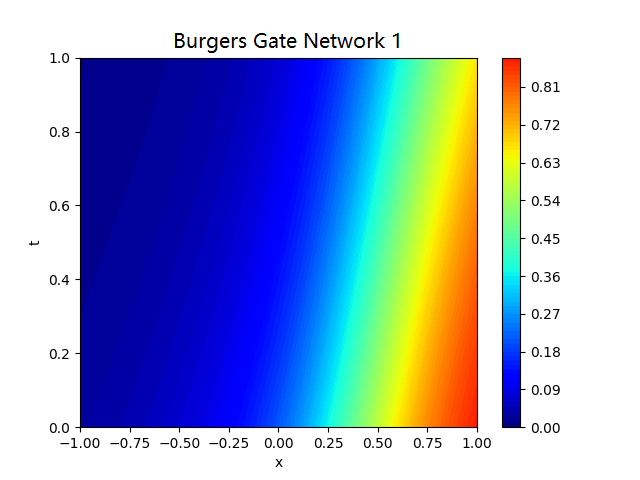}
\caption{The Burgers equation: Visualization of the gating network $G_1$ optimization trajectory via four snapshots, at epoch =  0, 1E4, 2E4, 3E4, from left to right and from top to bottom.
The value for the second subnet, $G_2$, is easily calculated using partition-of-unity property gating networks, i.e., $\sum_i G_i = 1$. }
\label{gif:burgers}
\end{figure}

\subsubsection{APINN}
To mimic the hard decomposition based on whether $x > 0$, we pretrain the gate net $G$ on the function $(G(x, t))_1 = 1 - (G(x, t))_2 = exp(x-1)$, so that the first sub-PINN focuses on where $x$ is larger and the second sub-PINN focuses on where $x$ is smaller.The corresponding model is named APINN-X.
In addition, we pretrain the gate net $G$ on $(G(x, t))_1 = 1 - (G(x, t))_2 = 0.8$ to mimic multi-level PINN (MPINN) \cite{anonymous2022multilevel}, where the first sub-net focuses on the majority part, while the second one is responsible for the minority part. The corresponding model is named APINN-M.
All networks have a width of 20. The numbers of layers in the gate network, sub-PINN networks, and shared network are 2, 4, and 3, respectively, with 3462 / 3543 parameters depending on whether the gate network is trainable.
All models are finetuned by the L-BFGS optimizer until convergence after Adam optimization.

\subsubsection{Results}The results for the Burgers equation are shown in Table \ref{tab:burgers}. The reported relative $L_2$ errors are averaged over 10 independent runs, which are the best $L_2$ errors among their whole optimization processes. The error plots of XPINNv1 and APINN-X are visualized in Figures \ref{fig:burgers2} left and right, respectively.
\begin{itemize}
\item XPINN performs much worse than PINN, due to the large error near the interface, where the steep region is located. 
\item APINN-X performs the best because its parameters are more flexible than those of PINN, and it does not require interface conditions like in XPINN, so it can model the steep region well.
\item APINN-M performs worse than APINN-X, which means that MPINN initialization is worse than the XPINN one in this Burgers problem.
\item APINN-X-F with a fixed gate function performs slightly worse than PINN and APINN, which justifies the flexibility of trainable domain decomposition. However, even without fine-tuning the domain decomposition, APINN-X-F can still outperform XPINN significantly, which shows the effectiveness of soft domain partition.
\end{itemize}

\subsubsection{Visualization of Gating Networks}
Some representative optimized gating networks after convergence are visualized in Figure \ref{fig:burgers_gate}. 
In the first row, we visualize two gate nets of APINN-X. Despite the fact that their optimized gates differ, they retain the original left-and-right decomposition with the change in interface position.Thus, their $L_2$ errors are similar.
In the second row, we show two gate nets of APINN-M. Their performances differ a lot, and they weight the two subnets differently. The third figure uses a weight $\approx 0.9$ for subnet-1 and a weight $\approx 0.1$ for subnet-2, while the fourth figure uses $\approx 0.6$ weight for subnet-1 and $\approx 0.4$ weight for subnet-2. It means that the training of MPINN-type decomposition is unstable, that APINN-M is worse than its XPINN counterpart in the Burgers problem, and that the weight in MPINN-type decomposition is crucial to its final performance.
From these examples, we can see that initialization is crucial for APINN's success. Despite the optimization, the trained gate will still be similar to the initialization.

Furthermore, we visualize the optimization trajectory of the gating network for the first subnet in the Burgers equation in Figure \ref{gif:burgers}, where each snapshot is the gating net at epoch = 0, 1E4, 2E4, 3E4. That for the second subnet $G_2$ can be easily computed using the property of partition-of-unity $G_1 + G_2 = 1$. The trajectory is smooth, and the gating net gradually converges by moving the interface from left to right and shifting the interface.



\begin{figure}
\centering
\includegraphics[scale=0.5]{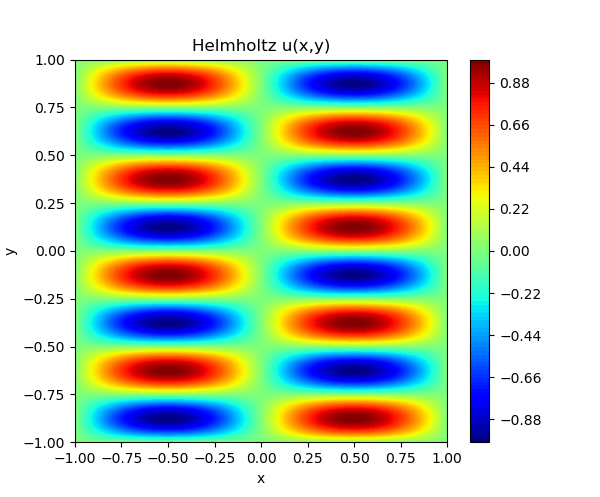}
\includegraphics[scale=0.5]{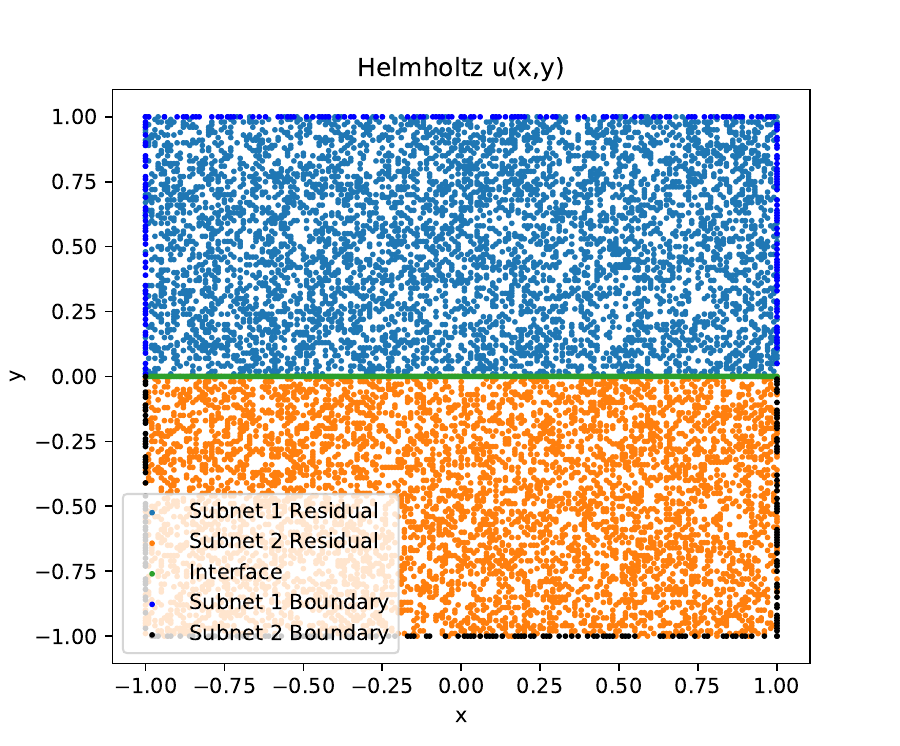}
\caption{The Helmholtz equation. Left: ground truth solution. Right: training points of XPINN.}
\label{fig:helmholtz1}
\end{figure}
\subsection{Helmholtz Equation}
Problems in physics including seismology, electromagnetic radiation, and acoustics are solved using the Helmholtz equation, which is given by
\begin{equation}
\begin{aligned}
& u_{xx} + u_{yy} + k^2 u  = q(x,y), x \in [-1,1], y\in [-1,1].\\
& u(-1,y) = u(1,y) = u(x,-1) = u(x,1) = 0.\\
& q(x, y) = \left(- (a_1 \pi)^2 - (a_2 \pi)^2 + k^2 \right)\sin(a_1\pi x)\sin(a_2\pi y).
\end{aligned}
\end{equation}
The analytic solution is 
\begin{equation}
u(x, y) = \sin(a_1\pi x)\sin(a_2\pi y),
\end{equation}
and is shown in Figure \ref{fig:helmholtz1} left.

In this case, XPINNv1 performs worse than PINN due to the large errors near the interface. With additional regularization, XPINNv2 reduces 47\% the relative $L_2$ error compared to PINN, but it still performs worse than our APINN due to the overfitting effect caused by the small availability of the training data in each sub-domain. 

\begin{table}[]
\centering
\caption{Results for the Helmholtz equation.}
\label{tab:helmholtz}
\begin{tabular}{|c|c|c|c|c|}
\hline
Model & PINN & XPINNv1 & XPINNv2 & -\\ \hline
Rel. $L_2$ & 2.438E-3$\pm$5.196E-4 & 5.222E-2$\pm$4.001E-3 & 1.297E-3$\pm$1.786E-4 & - \\ \hline
Model & APINN-X-F & APINN-X & APINN-M-F & APINN-M \\ \hline
Rel. $L_2$ & 1.554E-3$\pm$3.203E-4 & \textbf{1.275E-3$\pm$4.710E-4} & 1.911E-3$\pm$3.850E-4 & 1.477E-3$\pm$5.679E-4\\ \hline
\end{tabular}
\end{table}

\begin{figure}
\centering
\includegraphics[scale=0.8]{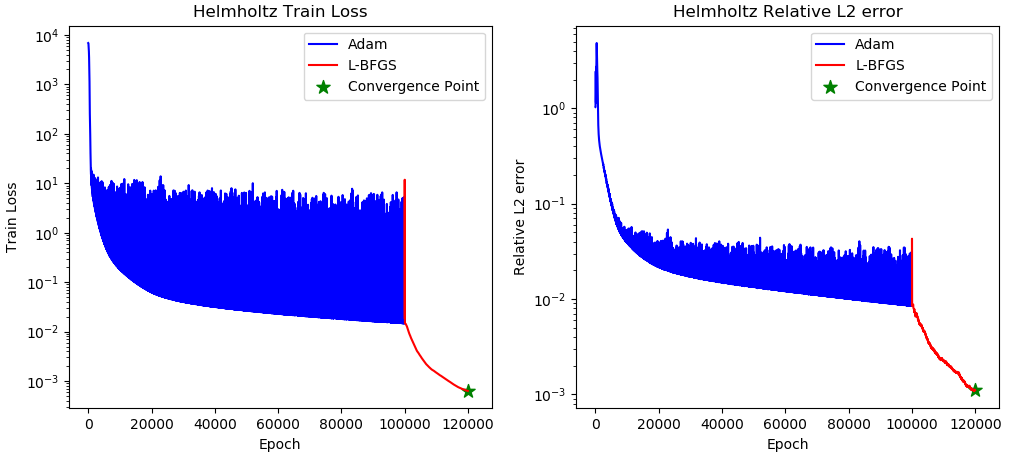}
\caption{The Helmholtz equation. Train loss and relative $L_2$ error during optimization. Blue: Adam optimization. Red: L-BFGS finetuning. Green: final convergence point. L-BFGS automatically stops since its convergence criterion is satisfied.}
\end{figure}

\subsubsection{PINN and Hard XPINN}
For PINN, we provide 400 boundary and 10000 residual points. 
The XPINN decomposes the domain based on whether $y > 0$, whose training points are shown in Figure \ref{fig:helmholtz1} right.
We provide 200 boundary points, 5000 residual points, and 400 interface points for the two sub-nets in XPINN.
Other settings of PINN and XPINN are the same as those in the Burgers equation.

\begin{figure}
\centering
\includegraphics[scale=0.35]{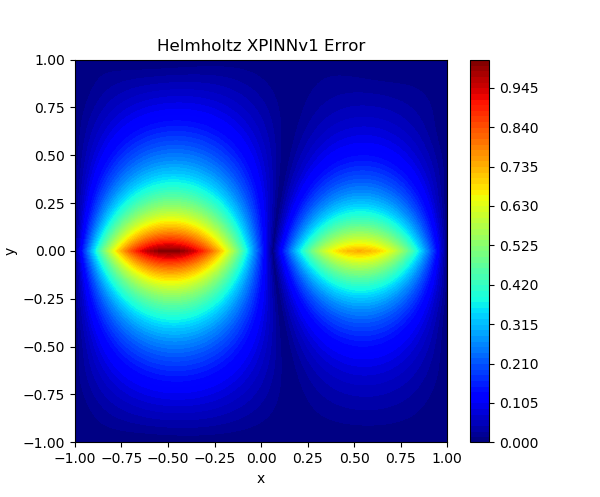}
\includegraphics[scale=0.35]{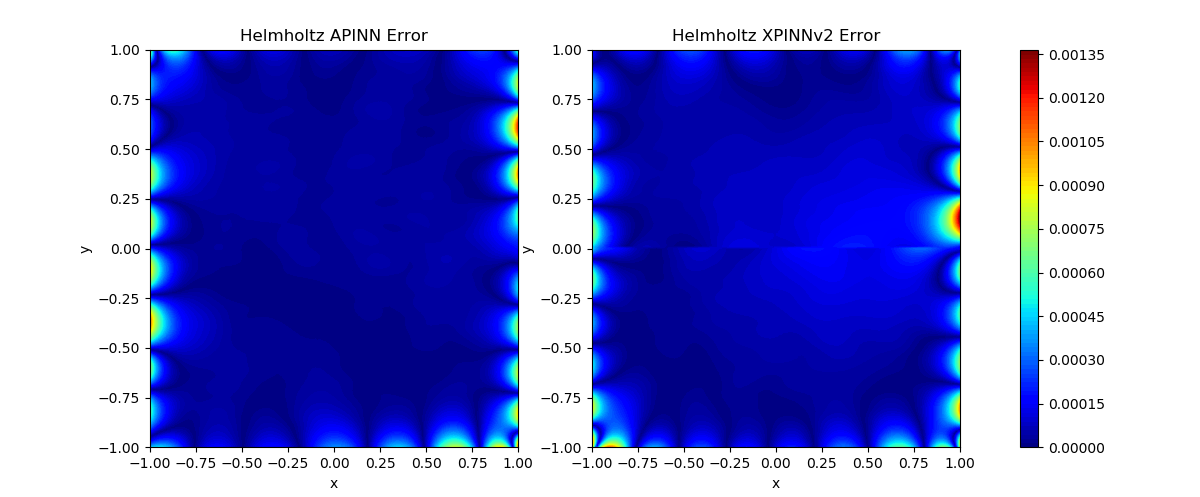}
\caption{{The Helmholtz equation. Left: error plot of XPINNv1. Middle: error plot of APINN. Right: error plot of XPINNv2.}} 
\label{fig:helmholtz2}
\end{figure}

\begin{figure}
\centering
\includegraphics[scale=0.7]{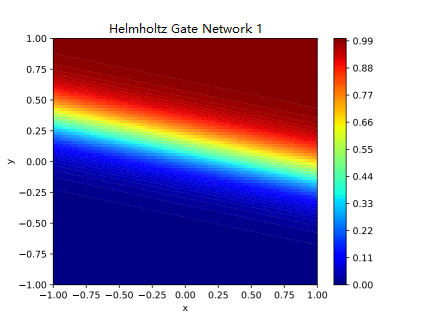}
\includegraphics[scale=0.7]{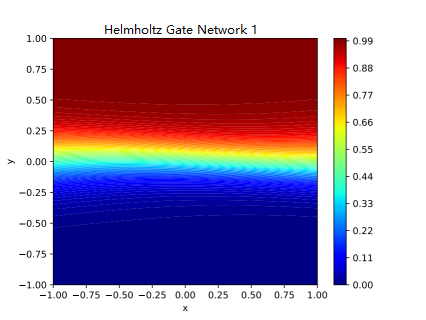}
\includegraphics[scale=0.7]{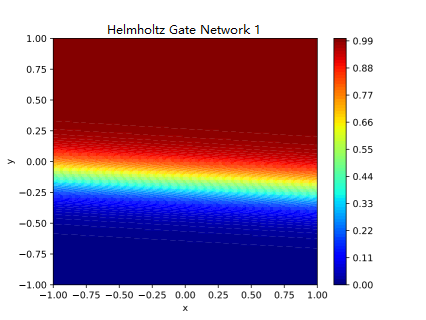}
\caption{The Helmholtz equation: APINN-X gate nets $G_1$ after convergence at the last epoch. Their relative $L_2$ errors are similar.}
\label{fig:helmholtz_gate}
\end{figure}

\subsubsection{APINN}
We pretrain the gate net $G$ on the function $(G(x, y))_1 = 1-(G(x, y))_2 = \exp(y-1)$ to mimic XPINN, and on $(G(x, y))_1 = 1-(G(x, y))_2 = 0.8$ to mimic MPINN. 
For other experimental settings, please refer to the introduction of APINN in the Burgers equation.



\subsubsection{Results}
The results for the Helmholtz equation are shown in Table \ref{tab:helmholtz}. The reported relative $L_2$ errors are averaged over 10 independent runs, which are selected as having the lowest errors during optimization. The error plots of XPINNv1, APINN-X and XPINNv2 are visualized in Figure \ref{fig:helmholtz2} left, middle, and right, respectively.
\begin{itemize}
\item XPINNv1 performs the worst, since its interface loss cannot enforce the interface continuity satisfactorily.
\item XPINNv2 performs significantly better than PINN, but it is worse than APINN-X, because it overfits in the two sub-domains a bit due to the small number of available training samples, compared with APINN-X. 
\item APINN-M performs worse than APINN-X due to bad initialization of the gating network.
\item The errors of APINN, XPINNv2 and PINN concentrate near the boundary, which is due to the gradient pathology \cite{wang2021understanding}.
\end{itemize}

\subsubsection{Visualization of Optimized Gating Networks}
The randomness of this problem is smaller, so that the final relative $L_2$ errors of different runs are similar. 
Some representative optimized gating networks after convergence of APINN-X are visualized in Figure \ref{fig:helmholtz_gate}. 
Specifically, every gating network maintains approximately the original decomposition into an upper and a lower domain, despite the fact that the interfaces change a bit in each run.
From these observations, the XPINN-type decomposition into an upper and a bottom domain is already satisfactory for XPINN. We also notice that the XPINN outperforms PINN, which is consistent with our observation.

\begin{figure}
\centering
\includegraphics[scale=0.45]{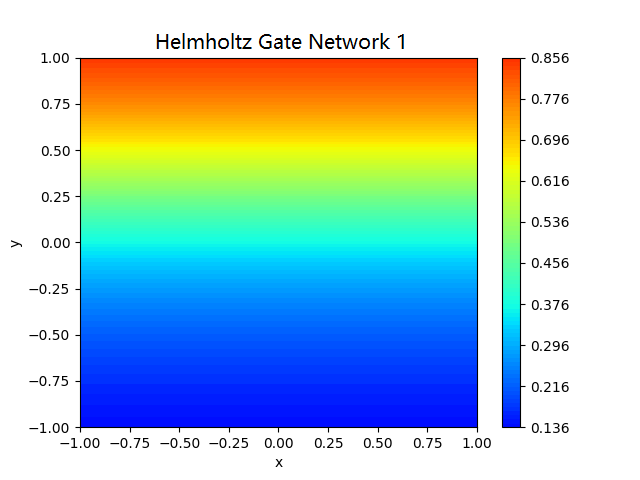}
\includegraphics[scale=0.45]{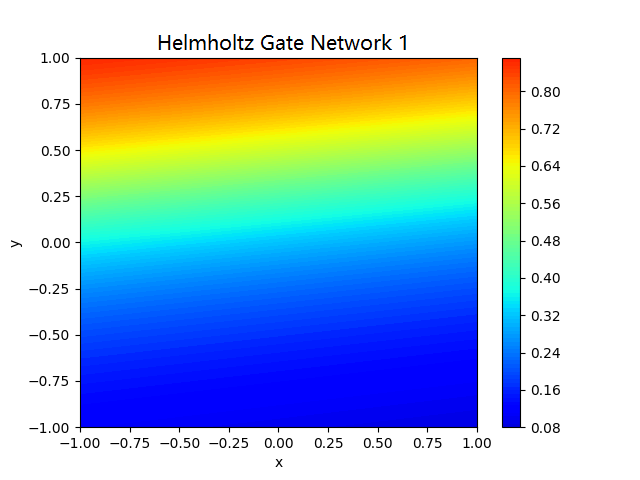}
\includegraphics[scale=0.45]{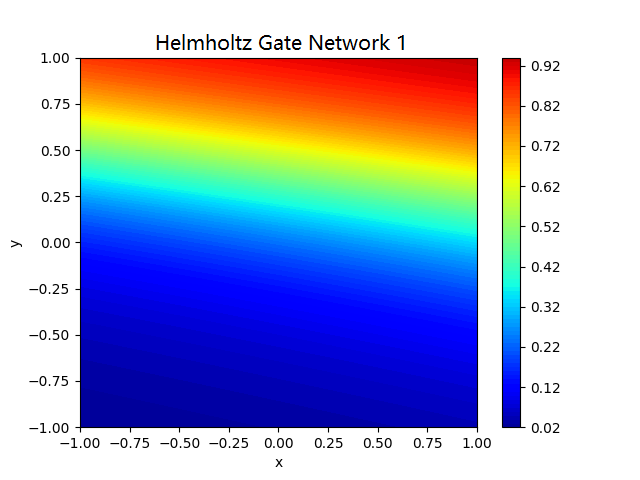}
\includegraphics[scale=0.45]{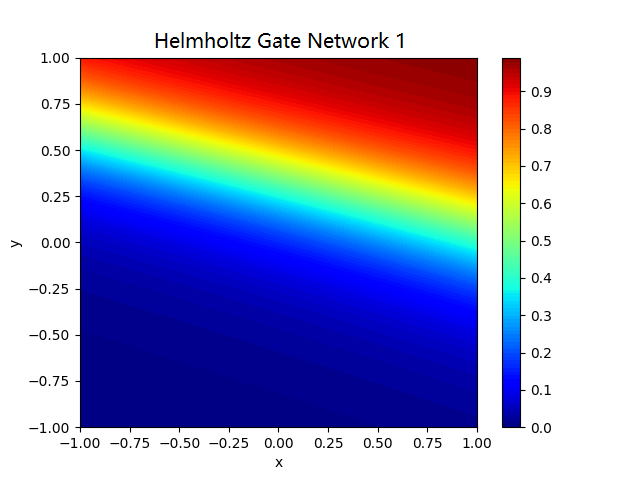}
\includegraphics[scale=0.45]{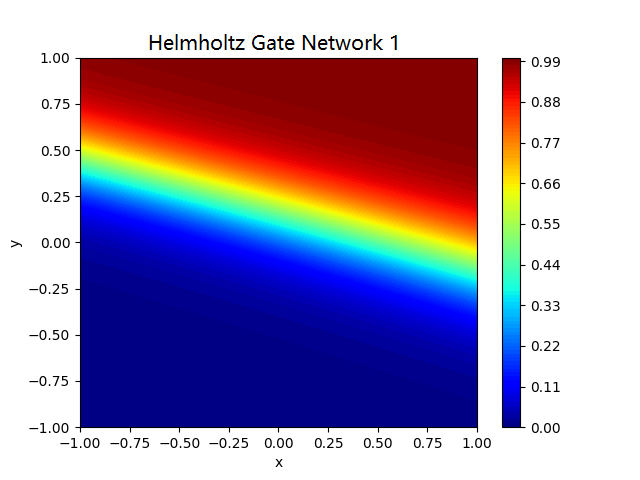}
\includegraphics[scale=0.45]{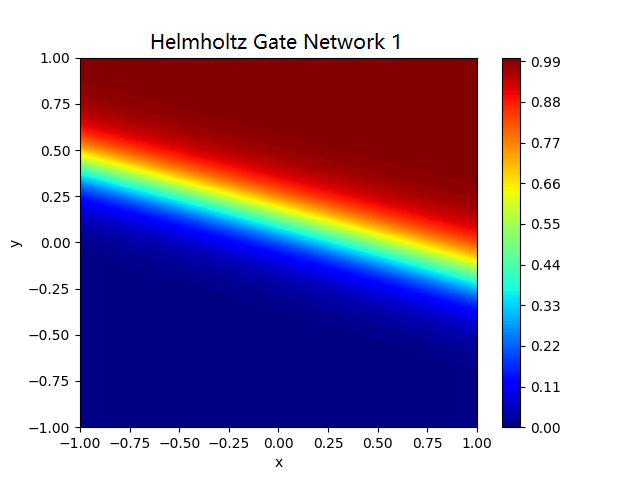}
\caption{The Helmholtz equation: visualization of the gating network $G_1$ optimization trajectory via six snapshots, at epoch =  0, 1E2, 2E2, 3E2, 4E2, and 5E2, from left to right and from top to bottom.
That for the second subnet $G_2$ can be easily computed using partition-of-unity property gating networks, i.e., $ \sum_i G_i = 1$.}
\label{gif:helmholtz}
\end{figure}

Furthermore, we visualize the optimization trajectory of the gating network for the first subnet in the Helmhotz equation in Figure \ref{gif:helmholtz}, where each snapshot is the gating net at epoch = 0 to 5E2 with 6 snapshots in all. That for the second subnet $G_2$ can be easily computed using partition-of-unity property gating networks, i.e., $ \sum_i G_i = 1$. The trajectory is similar to the case in the Burgers equation. Here, the gating net of the Helmhotz equation converges much faster than the one in the previous Burgers equation.

\begin{figure}
\centering
\includegraphics[scale=0.5]{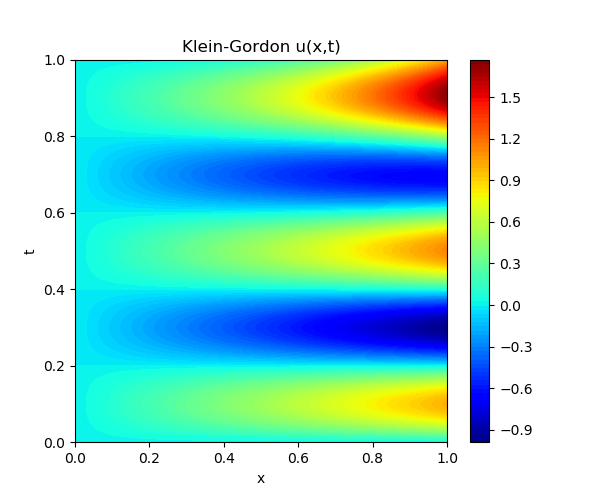}
\includegraphics[scale=0.5]{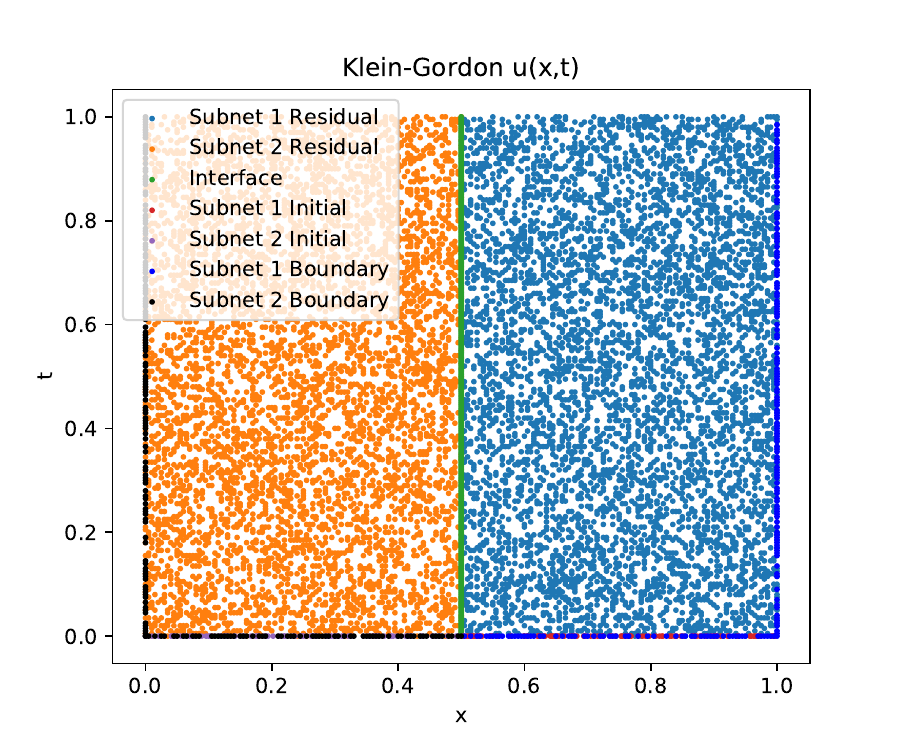}
\caption{The Klein-Gordon equation. Left: ground truth solution. Right: training points of XPINN.}
\label{fig:KG1}
\end{figure}
\begin{figure}
\centering
\includegraphics[scale=0.8]{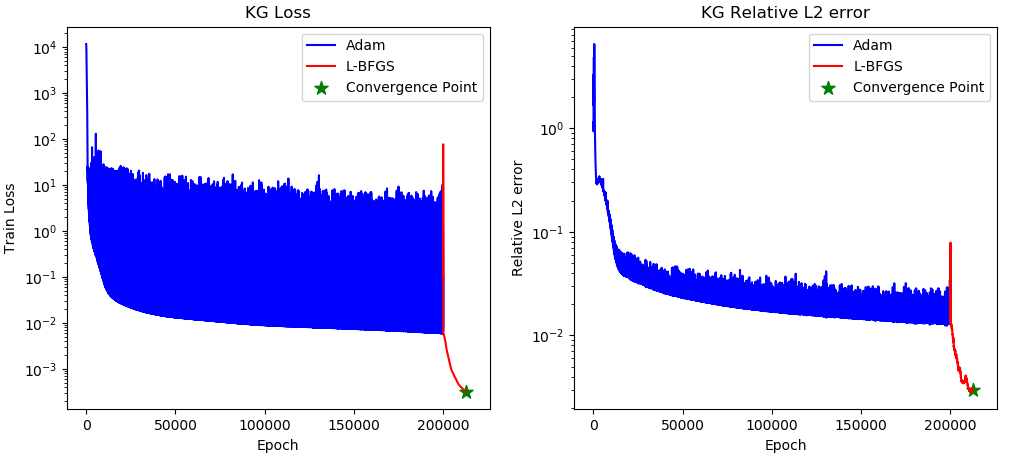}
\caption{The Klein-Gordon equation. Train loss and relative $L_2$ error during optimization. Green: final convergence point. L-BFGS automatically stops since its convergence criterion is satisfied.}
\end{figure}

\begin{table}[]
\centering
\caption{Results for the Klein-Gordon equation.}
\label{tab:KG}
\begin{tabular}{|c|c|c|c|c|}
\hline
Model & PINN & XPINNv1 & XPINNv2 & -\\ \hline
Rel. $L_2$ & 3.565E-3$\pm$9.412E-4 & 5.980E-1$\pm$7.601E-2 & 3.700E-3$\pm$2.741E-4 & - \\ \hline
Model & APINN-X-F & APINN-X & APINN-M-F & APINN-M\\ \hline
Rel. $L_2$ & 3.195E-3$\pm$8.112E-4 & {3.030E-3$\pm$1.474E-3} & 3.197E-2$\pm$6.253E-3 & \textbf{2.846E-3$\pm$8.568E-4}\\ \hline
\end{tabular}
\end{table}

\begin{figure}
\centering
\includegraphics[scale=0.36]{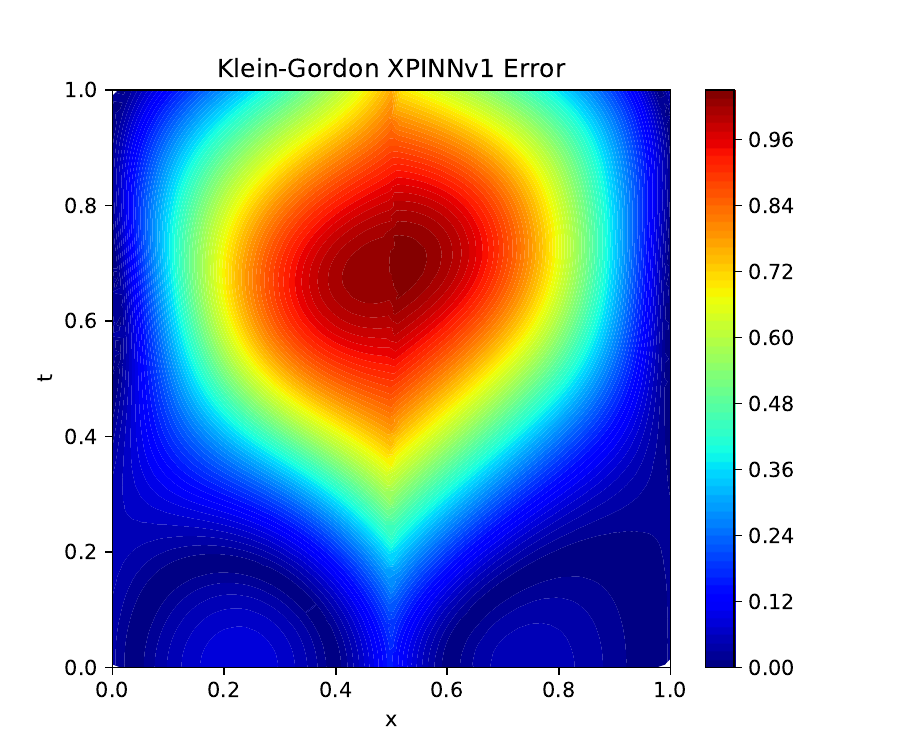}
\includegraphics[scale=0.5]{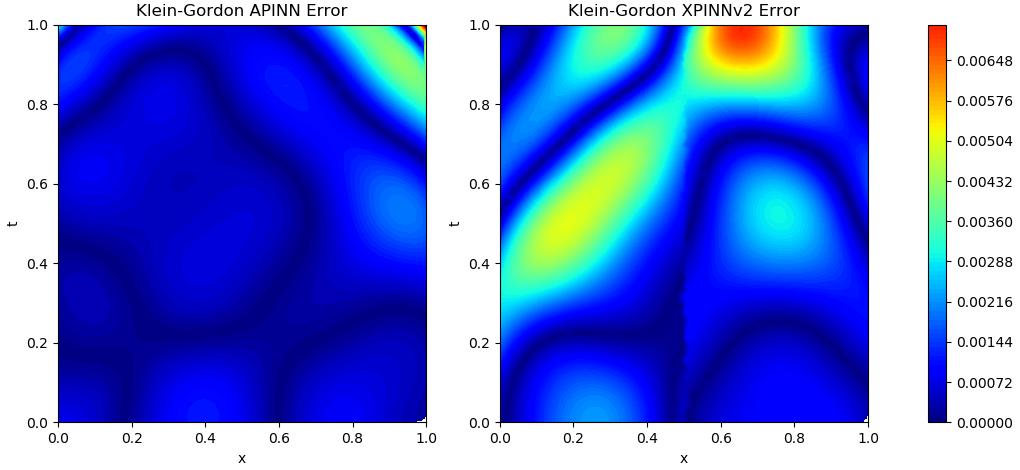}
\caption{{The Klein-Gordon equation. Left: error plot of XPINNv1. Middle: error plot of APINN-X. Right: error plot of XPINNv2. Since XPINNv1 exhibits large errors near the interface, it has its own colorbar, since errors of other models become negligible compared with XPINNv1. APINN and XPINNv2 share the same colorbar for clear comparison since their error scales are similar.}}
\label{fig:KG2}
\end{figure}
\subsection{Klein-Gordon Equation}
In modern physics, the equation is used in a wide variety of fields, such as particle physics, astrophysics, cosmology, classical mechanics, etc. and it is given by
\begin{equation}
\begin{aligned}
& u_{tt} - u_{xx} + u^3  = f(x,t), x \in [0,1], t\in [0,1].\\
& u(x,0) = u_t(x,0) = 0.\\
& u(x,t) = h(x,t), x\in\{0,1\}, t\in[0,1].
\end{aligned}
\end{equation}
Its boundary and initial conditions are given by the ground truth solution:
\begin{equation}
u(x, y) = x \cos(5 \pi t) + (xt)^3,
\end{equation}
and is shown in Figure \ref{fig:KG1} left. In this case, XPINNv1 performs worse than PINN due to the large errors near the interface induced by unsatisfactory continuity between sub-nets, while XPINNv2 performs similarly to PINN. APINN performs much better than XPINNv1 and better than PINN and XPINNv2.

\subsubsection{PINN and Hard XPINN}
The experimental settings of PINN and XPINN are identical to those of the previous Helmholtz equation, with the exception that XPINN now decomposes the domain based on whether $x > 0.5$ and Adam optimization is performed for 200k epochs. 


\subsubsection{APINN}
We pretrain the gate net $G$ on the function $(G(x, t))_1 = 1 - (G(x, t))_2 = \exp(-x)$ to mimic XPINN, and on $(G(x, t))_1 = 1- (G(x, t))_2 = 0.8$ to mimic MPINN. 
For other experimental settings, please refer to the introduction of APINN in the first equation.



\subsubsection{Results}
The results for the Klein-Gordon equation are shown in Table \ref{tab:KG}. The reported relative $L_2$ errors are averaged over 10 independent runs. The error plots of XPINNv1, APINN-X, and XPINNv2 are visualized in Figures \ref{fig:KG2} left, middle, and right, respectively.
\begin{itemize}
\item XPINNv1 performs the worst, since the interface loss of XPINNv1 cannot enforce the interface continuity well, while XPINNv2 performs similarly to PINN, since the two factors in XPINN generalization reach a balance. 
\item APINN performs better than all XPINNs and PINNs, and APINN-M is slightly better than APINN-X.
\end{itemize}

\begin{table}[]
\centering
\caption{Results for the Wave equation.}
\label{tab:wave}
\begin{tabular}{|c|c|c|c|c|}
\hline
Model & PINN & XPINNv2 & APINN-X & APINN-M\\ \hline
Rel. $L_2$ & 1.900E-3$\pm$3.375E-4 & 1.378E-3$\pm$2.424E-4 & {1.492E-3$\pm$7.041E-4} & \textbf{1.299E-3$\pm$2.941E-4} \\ \hline
\end{tabular}
\end{table}

\subsection{Wave Equation}
\begin{figure}
\centering
\includegraphics[scale=0.5]{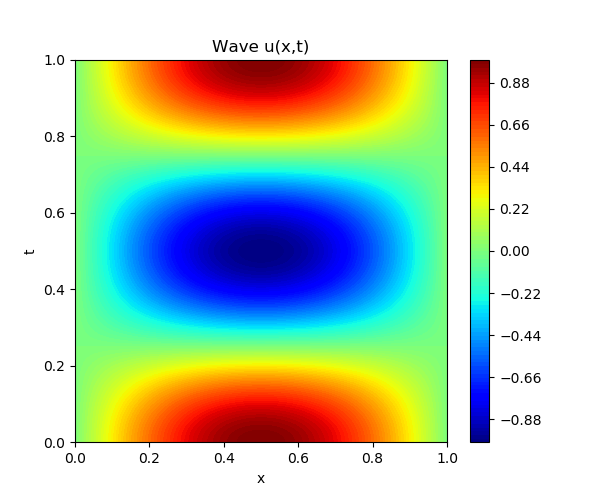}
\includegraphics[scale=0.5]{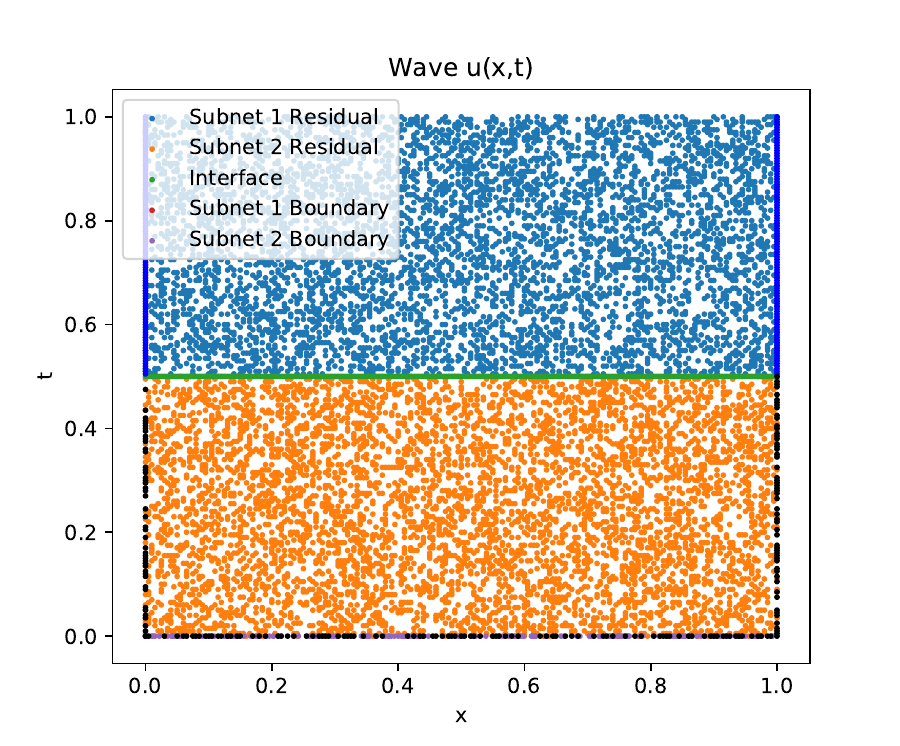}
\caption{The Wave equation. Left: ground truth solution. Right: training points of XPINN.}
\label{fig:wave1}
\end{figure}

We consider a wave problem given by
\begin{equation}
\begin{aligned}
& u_{tt} = 4 u_{xx}, x \in [0,1], t\in [0,1].
\end{aligned}
\end{equation}
The boundary and initial conditions are given by the ground truth solution:
\begin{equation}
u(x, t) = \sin(\pi x) \cos(2 \pi t),
\end{equation}
and is shown in Figure \ref{fig:wave1} left.

In this example, XPINN is already significantly better than PINN because its relative $L_2$ error is 27
However, APINN still performs slightly better than XPINN, even if XPINN is already good enough.

\begin{figure}
\centering
\includegraphics[scale=0.8]{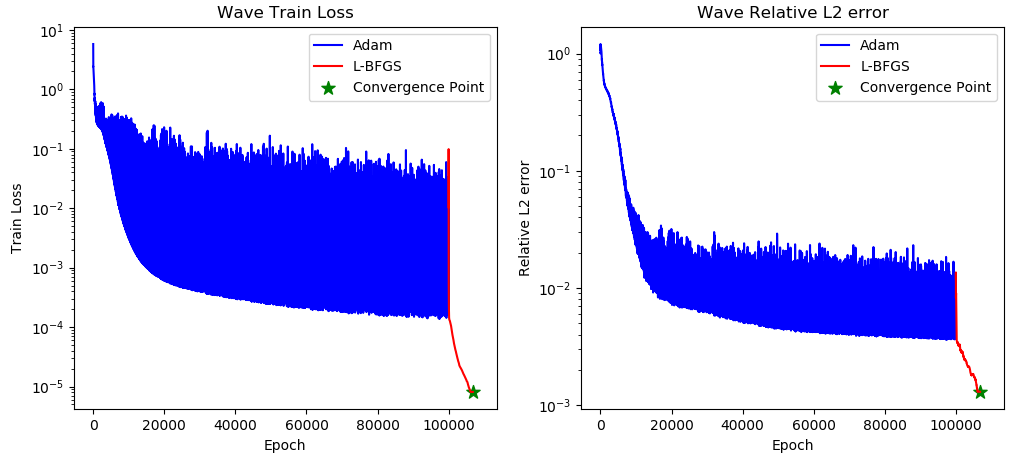}
\caption{The Wave equation. Train loss and relative $L_2$ error during optimization. Blue: Adam optimization. Red: L-BFGS finetuning. Green: final convergence point. L-BFGS automatically stops since its convergence criterion is satisfied.}
\end{figure}

\begin{figure}
\includegraphics[scale=0.6]{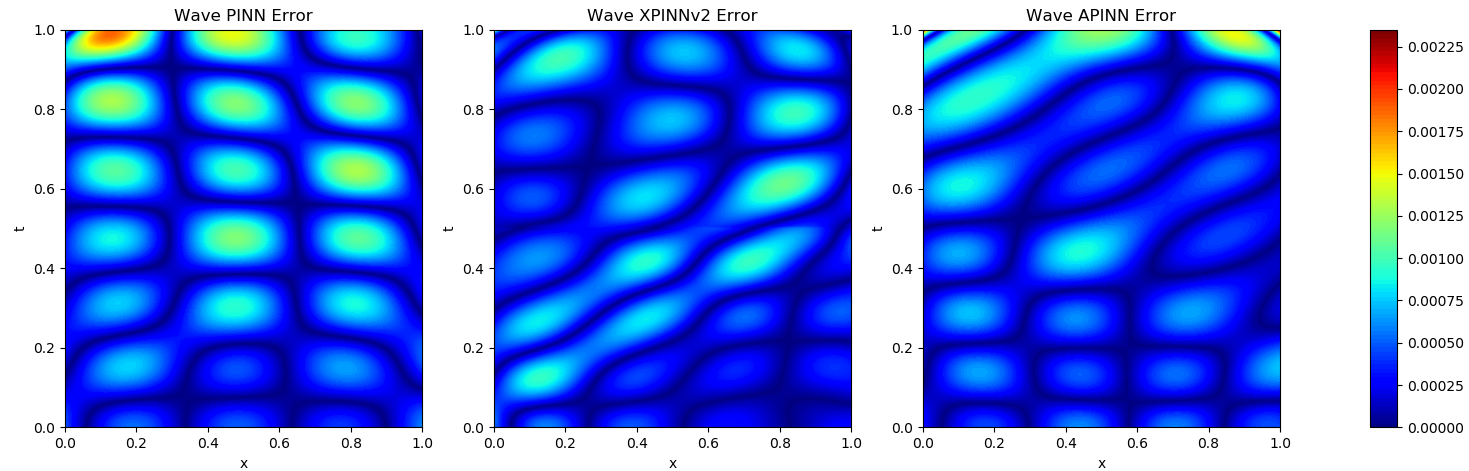}
\caption{{The Wave equation. Left: error plot of PINN. Middle: error plot of XPINNv2. Right: error plot of APINN.}}
\label{fig:wave2}
\end{figure}

\subsubsection{PINN and Hard XPINN}
We use a 10-layer tanh network with 3441 neurons and 400 boundary points and 10,000 residual points for PINN.We use 20 weight on the boundary and unity weight for the residual. We train PINN using the Adam optimizer for 100k epochs at an 8E-4 learning rate.
XPINN decomposes the domain based on whether $t > 0.5$.
The weights for boundary loss, residual loss, interface boundary loss, interface residual loss, and interface first-order derivative continuity loss are 20, 1, 20, 0, 1, respectively. The sub-nets are 6-layer tanh networks of 20-width with 3522 neurons in total, and we provide 200 boundary points, 5000 residual points, and 400 interface points for all sub-nets in XPINN. The training points of XPINN are visualized in Figure \ref{fig:wave1} right. We train XPINN using the Adam optimizer for 100k epochs at a learning rate of 1e-4. 

\subsubsection{APINN}
The APINNs mimic XPINN by pretraining on $(G(x, t))_1 = 1-(G(x, t))_2= \exp(-t)$ and mimic MPINN by pretraining on $(G(x, t))_1 =1-(G(x, t))_2= 0.8$. 
For other experimental settings, please refer to the introduction of APINN in the first equation.


\begin{figure}
\centering
\includegraphics[scale=1]{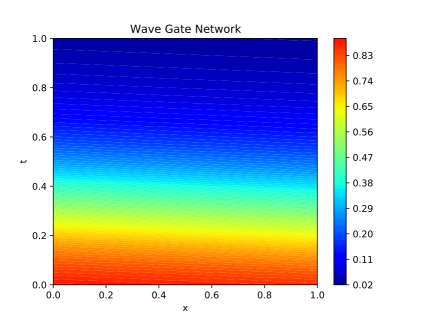}
\includegraphics[scale=1]{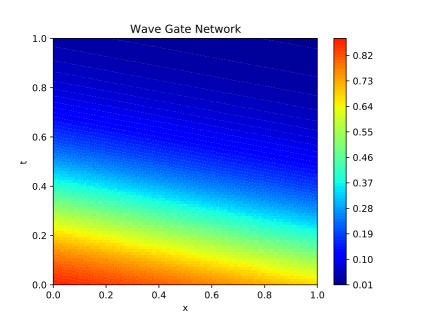}
\includegraphics[scale=1]{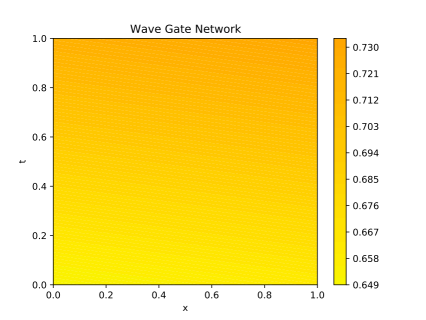}
\includegraphics[scale=1]{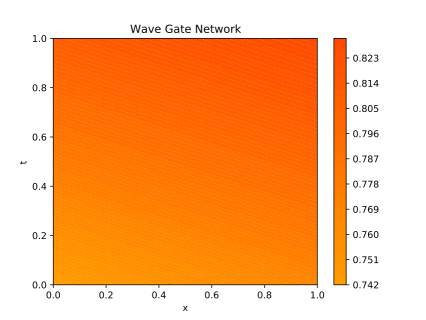}
\caption{The Wave equation: APINN gate nets $G_1$ after convergence at the last epoch. 
That for the second subnet $G_2$ can be easily computed using partition-of-unity property gating networks, i.e., $ \sum_i G_i = 1 $.
First row: those of APINN-X with two different random seeds. Relative $L_2$ errors = 1.477E-3, 1.527E-3.
Second row: those of APINN-M with two different random seeds. Relative $L_2$ errors = 1.055E-3, 1.315E-3.}
\label{fig:wave_gate}
\end{figure}

\subsubsection{Results}
The results for the wave equation are shown in Table \ref{tab:wave}. The reported relative $L_2$ errors are averaged over 10 independent runs, which are selected as the error at the epoch with the smaller training loss among the last 10\% epochs. The error plots of PINN, XPINNv2, and APINN-X are visualized in Figures \ref{fig:wave2} left, middle, and right, respectively.
\begin{itemize}
\item Although XPINN is already much better than PINN and reduces by 27\% the relative $L_2$ of PINN, APINN can still slightly improve over XPINN and performs the best among all models. In particular, APINN-M outperforms APINN-X.
\end{itemize}

\subsubsection{Visualization of Optimized Gating Networks}
Some representative optimized gating networks after convergence are visualized in Figure \ref{fig:wave_gate}. The first row shows the gate networks of optimized APINN-X, while the second row shows those of APINN-M.
In this case, the variance is much smaller, and the optimized gate nets maintain the characteristics at initialization, i.e., those of APINN-X remain an upper-and-lower decomposition and those of APINN-M remain a multi-level partition. Gate nets under the same initialization are also similar in different independent runs, which is consistent with their similar performances.

\subsection{Boussinesq-Burger Equation}

\begin{figure}
\centering
\includegraphics[scale=0.35]{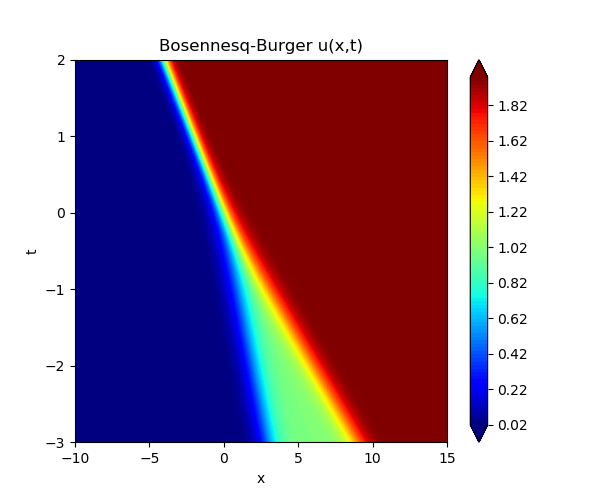}
\includegraphics[scale=0.35]{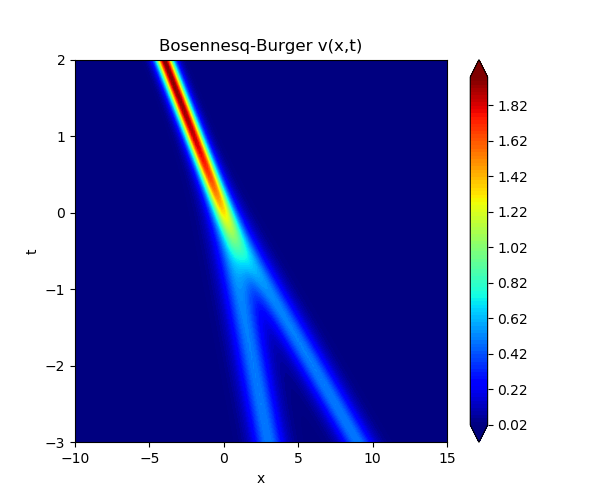}
\includegraphics[scale=0.35]{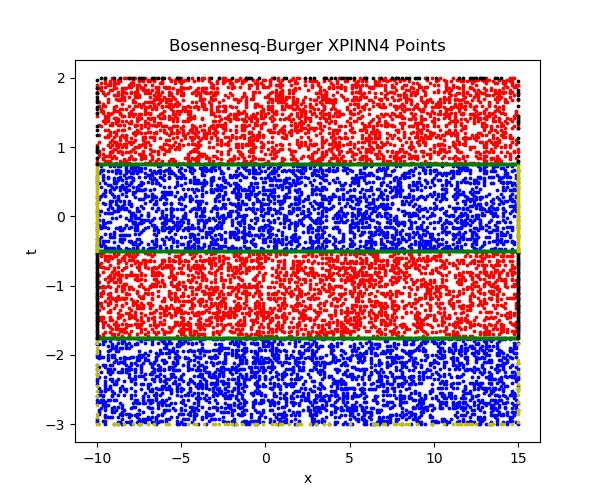}
\caption{The Boussinesq-Burger equation. Left and middle: ground truth solution for $u$ and $v$. Right: training points of XPINN4 with four subdomains.}
\label{fig:BB}
\end{figure}

Here we consider the Boussinesq-Burger system, which is a nonlinear water wave model consisting of two unknowns. A thorough understanding of such a model's solutions is important in order to apply it to harbor and coastal designs.
The Boussinesq-Burger equation under consideration is given by
\begin{equation}
\begin{aligned}
u_t = 2 u u_x + \frac{1}{2}v_x, \quad v_t = \frac{1}{2}v_{xxx} + 2 (uv)_x, \quad x \in [-10,15], t \in [-3,2],
\end{aligned}
\end{equation}
where the Dirichlet boundary condition and the ground truth solution is given in \cite{lin2022two}, and shown in Figure \ref{fig:BB} (left and middle) for the unknown $u$ and $v$, respectively.
In this experiment, we consider a system of PDEs, and try XPINN and APINN with more than two subdomains.
\begin{figure}
\centering
\includegraphics[scale=0.8]{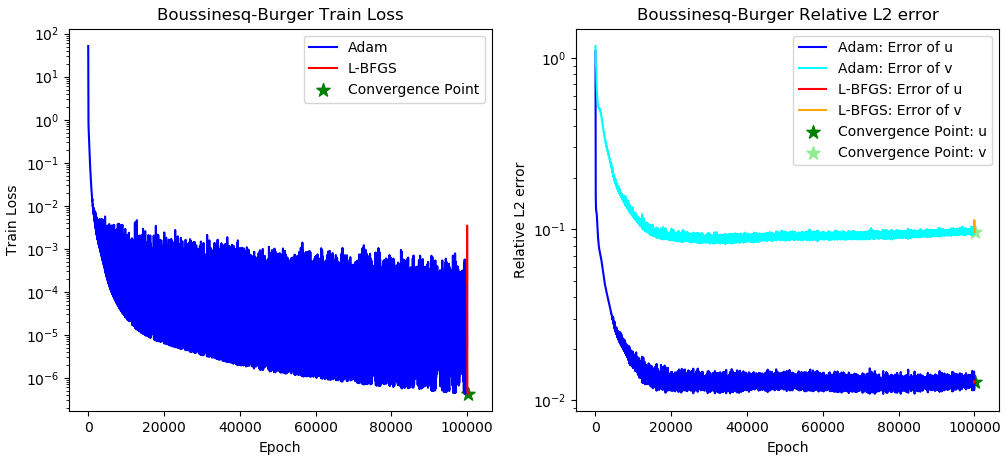}
\caption{The Boussinesq-Burger equation. Train loss and relative $L_2$ error during optimization. In this case, Adam can already train the model to convergence, so additional L-BFGS converges fast due to its stopping criterion.}
\end{figure}

\subsubsection{PINN and Hard XPINN}
For PINN, we use a 10-layer Tanh network, and provide 400 boundary points and 10,000 residual points. We use 20 weight on the boundary and unity weight for the residual. It is trained by Adam \cite{kingma2014adam} with an 8E-4 learning rate for 100K epochs.

For domain decomposition of (hard) XPINN, we design two different strategies.
First, a XPINN with two subdomains decomposes the domain based on whether $t > -0.5$.
The sub-nets are 6-layer tanh networks of 20-width, and we provide 200 boundary points and 5000 residual points for every sub-net in XPINN. 
Second, a XPINN4 with four subdomains decomposes the domain based on $t = -1.75, -0.5, $and $0.75$ into 4 subdomains, whose training points are visualized in Figure \ref{fig:BB} right.
The sub-nets in XPINN4 are 4-layer tanh networks of 20-width, and we provide 100 boundary points and 2500 residual points for every sub-net in XPINN4.
The number of interface points is 400.
The weights for boundary loss, residual loss, interface boundary loss, interface residual loss, and interface first-order derivative continuity loss are 20, 1, 20, 0, 1, respectively.
We use the Adam optimizer to train XPINN and XPINN4 for 100k epochs with an 8E-4 learning rate.
To make a fair comparison, the parameter counts in PINN, XPINN, and XPINN4 are 6882, 7044, and 7368, respectively. 

\begin{figure}
\centering
\includegraphics[scale=0.26]{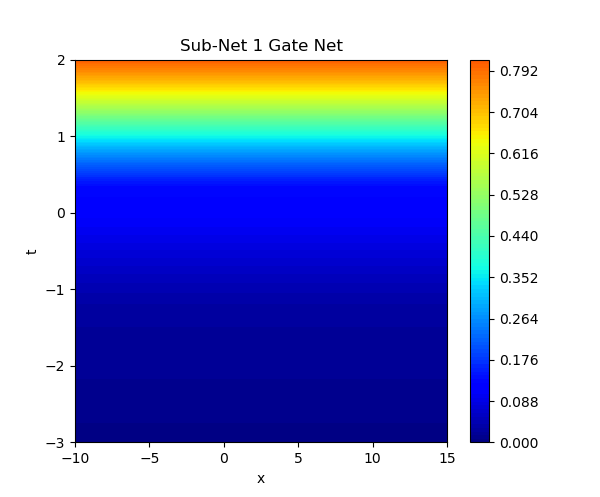}
\includegraphics[scale=0.26]{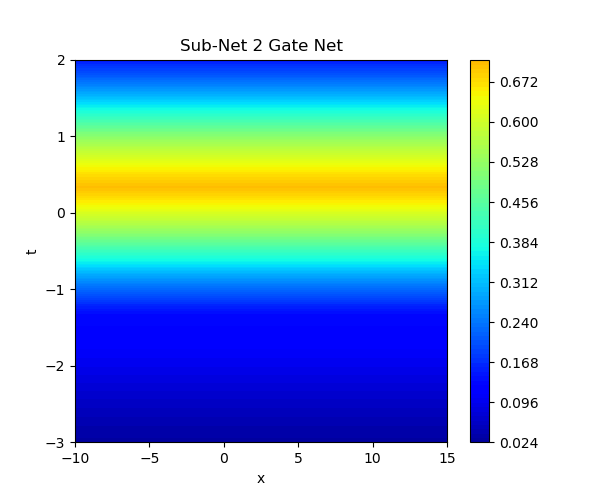}
\includegraphics[scale=0.26]{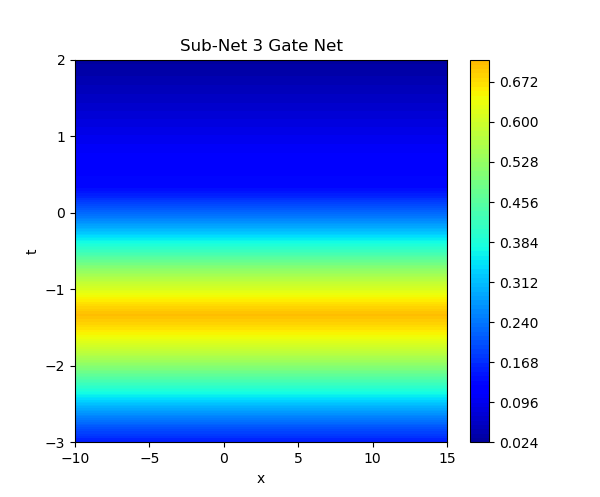}
\includegraphics[scale=0.26]{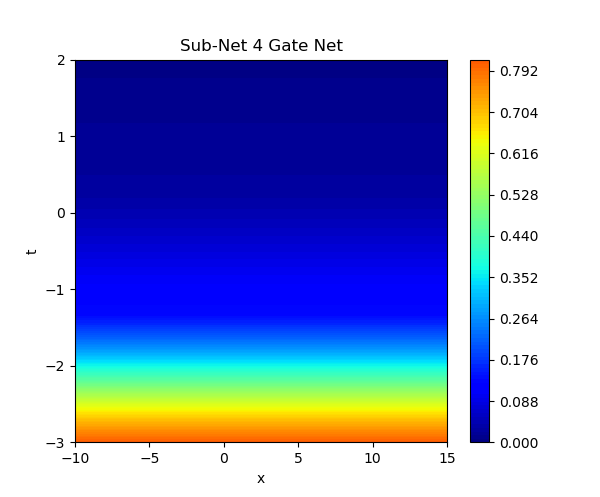}
\caption{The Boussinesq-Burger equation. APINN4-X pretrained gate nets, with four-dimensional output for weighted averaging the four subnets.}
\label{fig:BB_pretrained_gate}
\end{figure}

\subsubsection{APINN}
For APINN with two subdomains, we pretrain the gate net $G$ of APINN-X on the function $(G(x, t))_1 = 1- (G(x, t))_2= \exp(0.35 * (t - 2))$ to mimic XPINN, and pretrain that of APINN-M on the function $(G(x, t))_1 = 1-(G(x, t))_2 = 0.8$ to mimic MPINN.
In APINN-X and APINN-M, all networks have a width of 20. The numbers of layers in the gate network, sub-PINN networks, and shared network are 2, 4, and 5, respectively, with 6945 parameters in total.
For APINN with four subdomains, we pretrain the gate net $G$ of APINN4-X on the function $(G(x, t))_i = u_i(x,t) / (\sum_{i=1}^4 u_i(x,t))$, where $u_1(x,t)=\exp(t-2), u_2(x,t)=\exp(-|t - \frac{1}{3}|), u_3(x,t)=\exp(-|t + \frac{4}{3}|), u_4(x,t)=\exp(-3 - t)$, to mimic XPINN.
Furthermore, we pretrain that of APINN4-M on the function $(G(x, t))_1 = 0.8$, and $(G(x, t))_{2,3,4}=\frac{1}{15}$, to mimic MPINN.
The pretrained gate functions of APINN4-X are visualized in Figure \ref{fig:BB_pretrained_gate}.
In APINN4-X and APINN4-M, $h$ and $G$ are width 20, while $E_i$ is width 18. The numbers of layers in the gate network, sub-PINN networks, and the shared network
are 2, 4, and 3, respectively, with 7046 parameters in total.

\begin{table}[]
\centering
\caption{Relative $L_2$ error for the function $u$ in the Boussinesq-Burger equation.}
\label{tab:BB_u}
\begin{tabular}{|c|c|c|c|c|}
\hline
Model      & PINN    & XPINN   & XPINN4   & /        \\ \hline
Rel. $L_2$ & 1.470E-02$\pm$6.297E-03  & 1.456E-02$\pm$6.391E-03&3.254E-02$\pm$1.025E-02

  & /        \\ \hline
Model      & APINN-M & APINN-X & APINN4-M & APINN4-X \\ \hline
Rel. $L_2$ &  \bf1.091E-02$\pm$4.588E-03 & 1.388E-02$\pm$4.310E-03 & 1.328E-02$\pm$8.099E-03
 &  2.559E-02$\pm$6.554E-03
\\ \hline
\end{tabular}
\end{table}

\begin{table}[]
\centering
\caption{Relative $L_2$ error for the function $v$ in the Boussinesq-Burger equation.}
\label{tab:BB_v}
\begin{tabular}{|c|c|c|c|c|}
\hline
Model      & PINN    & XPINN   & XPINN4   & /        \\ \hline
Rel. $L_2$ & 1.106E-01$\pm$4.498E-02   &    9.786E-02$\pm$3.485E-02
 &  2.706E-01$\pm$9.078E-02
 & /        \\ \hline
Model      & APINN-M & APINN-X & APINN4-M & APINN4-X \\ \hline
Rel. $L_2$ & \bf8.185E-02$\pm$2.973E-02 &9.623E-02$\pm$2.446E-02 &9.616E-02$\pm$5.397E-02
& 1.676E-01$\pm$4.946E-02
 \\ \hline
\end{tabular}
\end{table}

\begin{figure}
\centering
\includegraphics[scale=0.75]{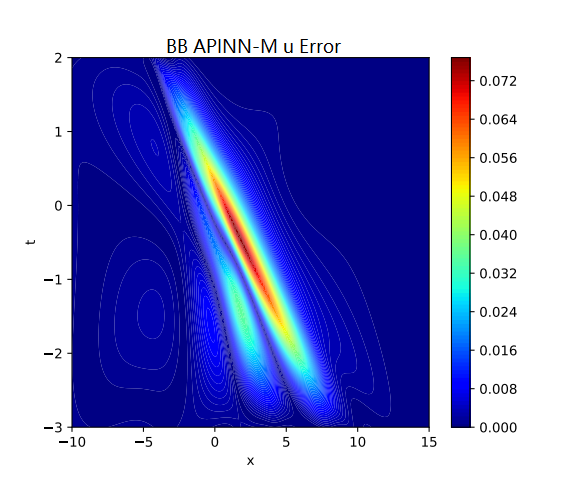}
\includegraphics[scale=0.75]{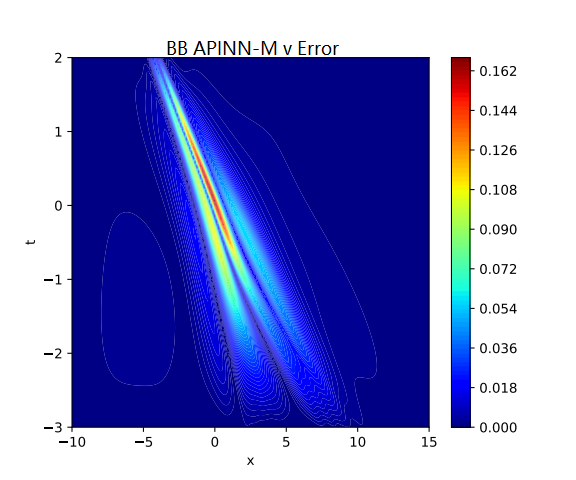}
\caption{The Boussinesq-Burger equation: Error of APINN-M.}
\label{fig:BB_Error}
\end{figure}

\begin{figure}
\centering
\includegraphics[scale=0.75]{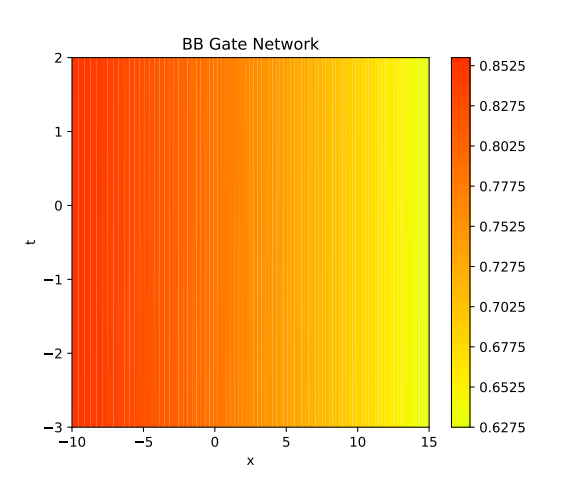}
\includegraphics[scale=0.75]{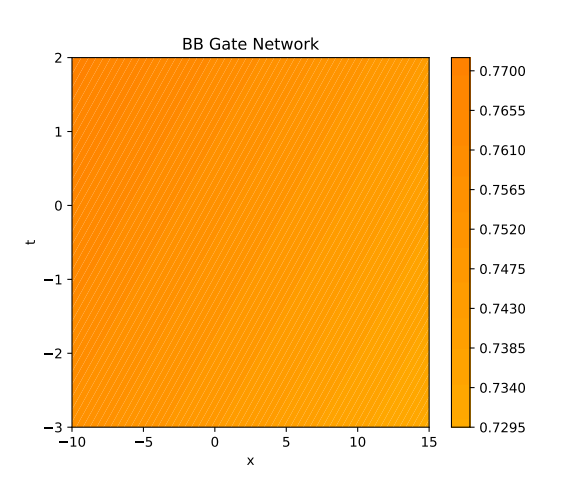}
\includegraphics[scale=0.7]{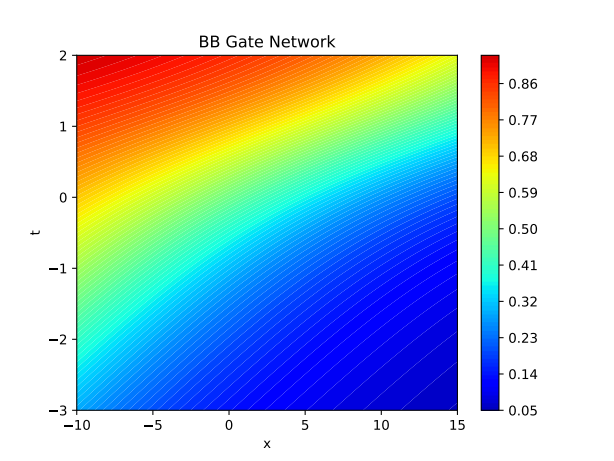}
\includegraphics[scale=0.7]{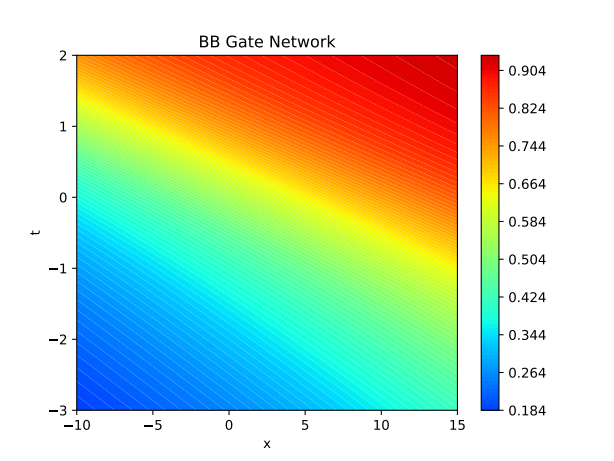}
\caption{The Boussinesq-Burger equation: visualization of trained gate networks $G_1$ of APINN-M (first row) and APINN-X (second row), after convergence, with two different random seeds for each model. Their relative $L_2$ errors are similar for the same type of APINNs.
}
\label{fig:BB_gate}
\end{figure}

\begin{figure}
\centering
\includegraphics[scale=0.23]{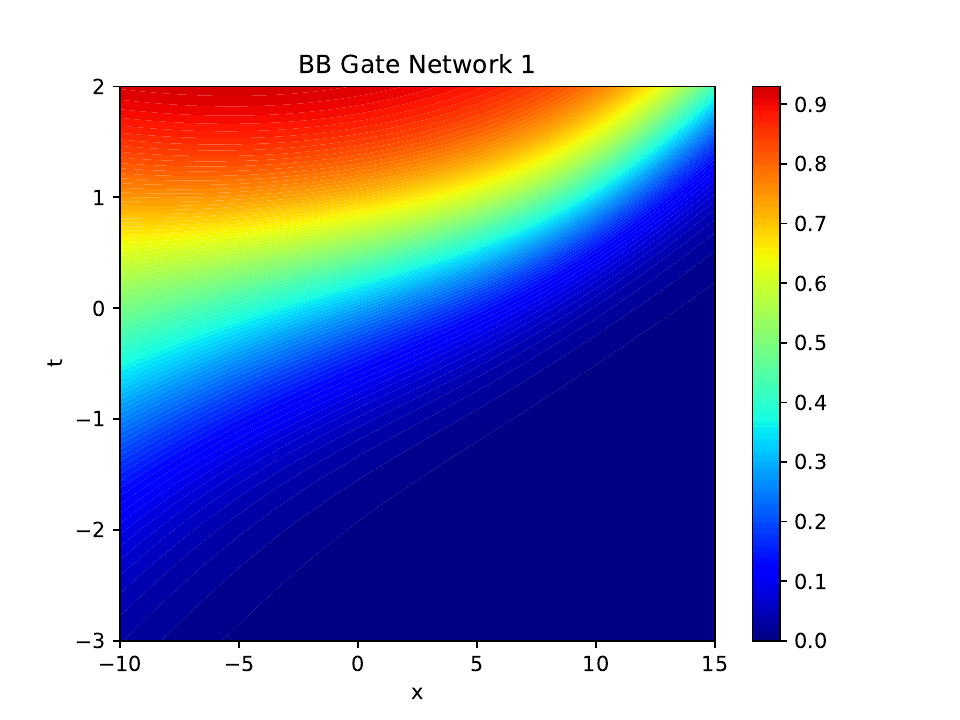}
\includegraphics[scale=0.23]{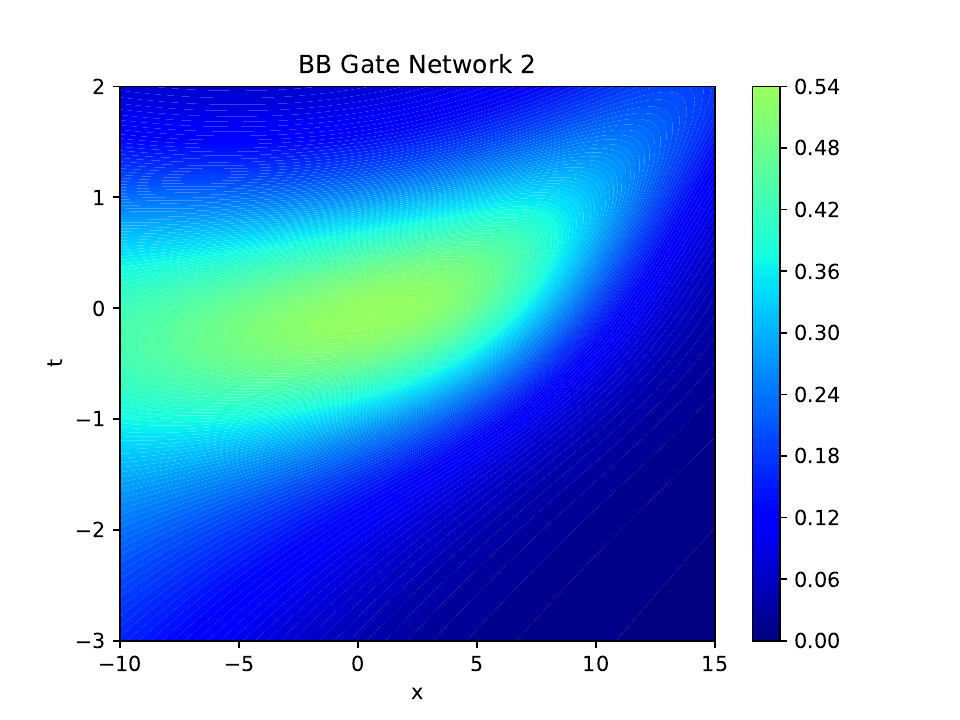}
\includegraphics[scale=0.23]{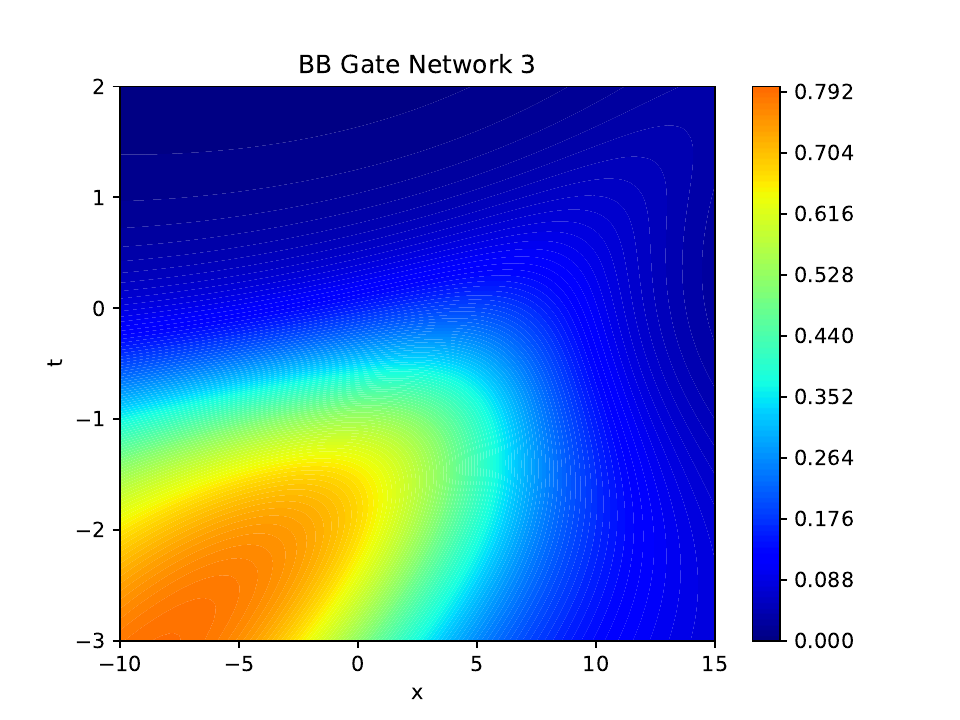}
\includegraphics[scale=0.23]{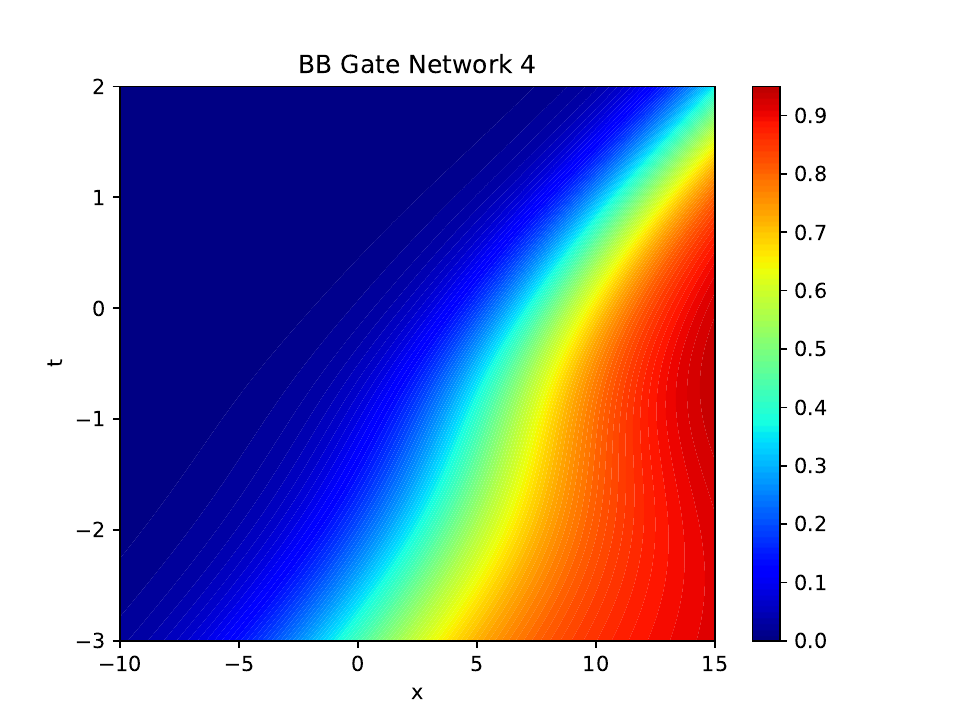}
\caption{The Boussinesq-Burger equation: Visualization of the four trained gate networks $G(x,t)_{1,2,3,4}$ in APINN4-X in one independent run, corresponding to the four subnets.}
\label{fig:BB4_gate_X}
\end{figure}

\begin{figure}
\centering
\includegraphics[scale=0.25]{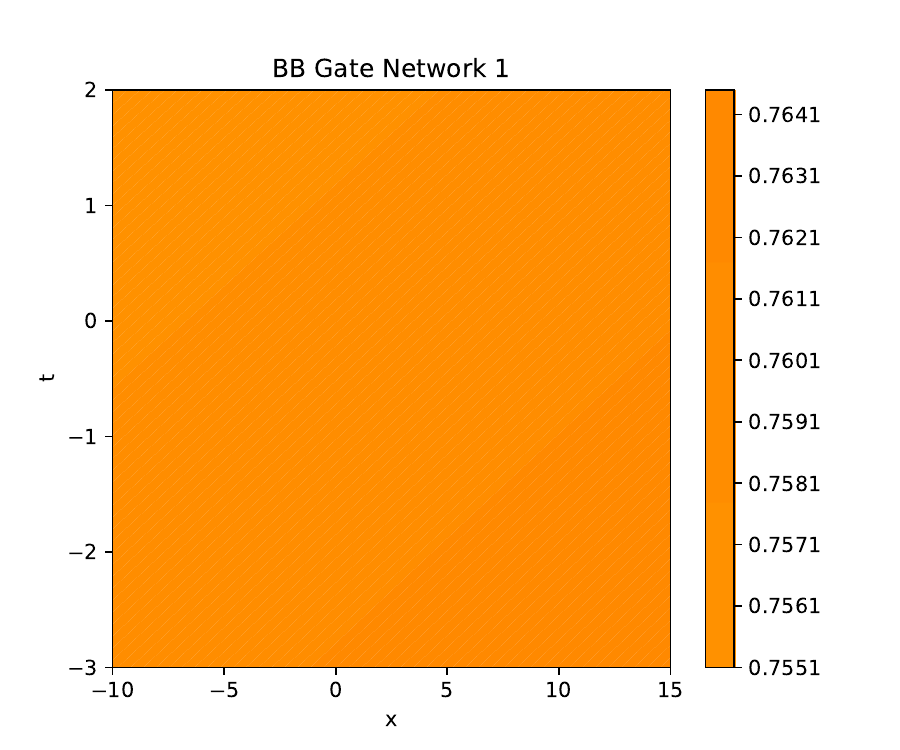}
\includegraphics[scale=0.25]{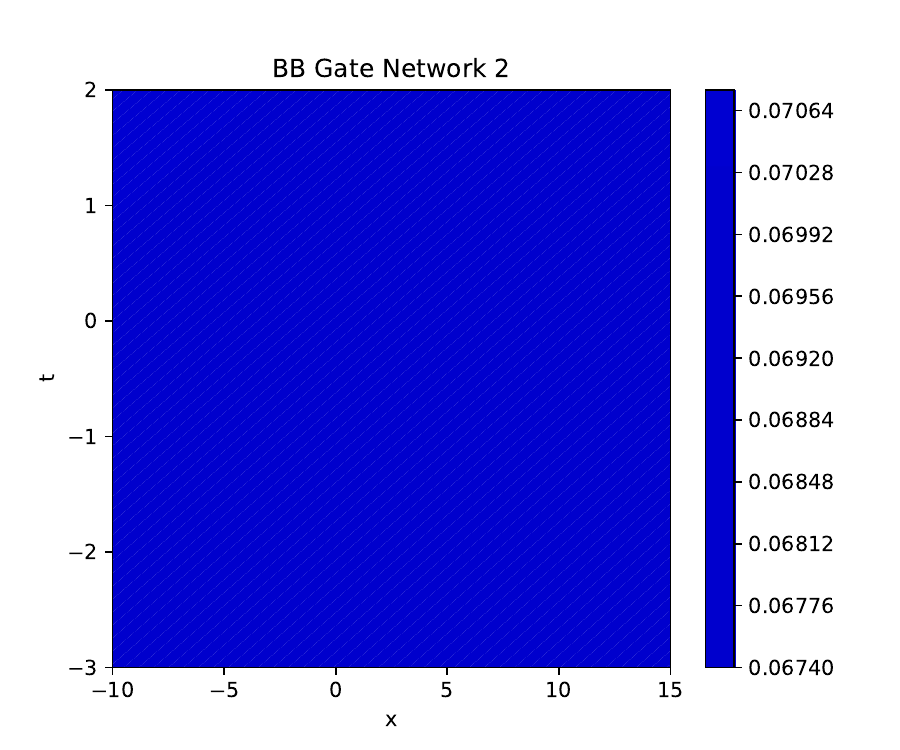}
\includegraphics[scale=0.25]{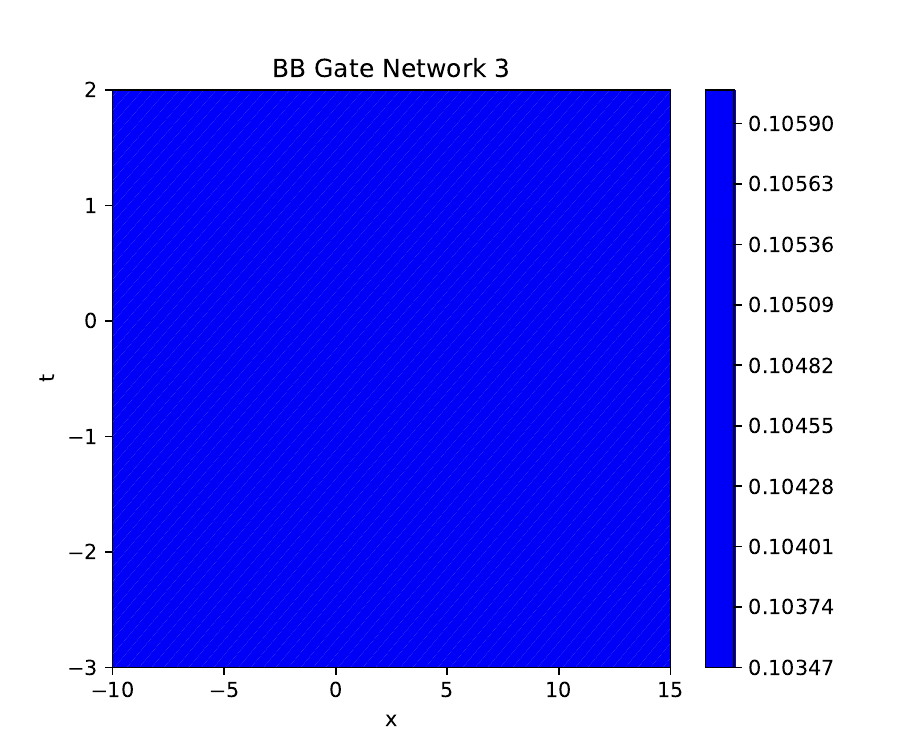}
\includegraphics[scale=0.25]{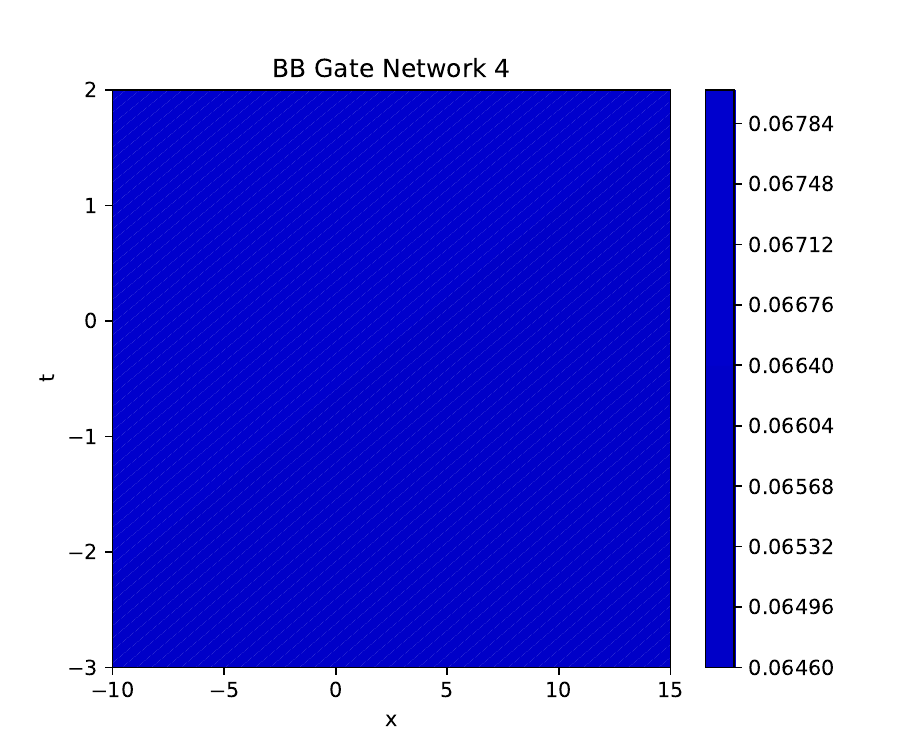}
\caption{The Boussinesq-Burger equation: Visualization of the four trained gate networks $G(x,t)_{1,2,3,4}$ in APINN4-M in one independent run, corresponding to the four subnets.}
\label{fig:BB4_gate_M}
\end{figure}

\subsubsection{Results}
The results for the Boussinesq-Burger equation are shown in Tables \ref{tab:BB_u} and \ref{tab:BB_v}. The reported relative $L_2$ errors are averaged over 10 independent runs, which are selected as the error at the epoch with the smaller training loss among the last 10\% epochs. The key observations are as follows.
\begin{itemize}
\item APINN-M performs the best.
\item APINN and XPINN with four sub-nets do not perform as well as their two sub-net counterparts, which may be due to the tradeoff between target function complexity and number of training samples in XPINN generalization. Also, more subdomains do not necessarily contribute to parameter efficiency.
\item The error of the best performing APINN-M is visualized in Figure \ref{fig:BB_Error}, which is concentrated near the steep regions, where the solution changes rapidly.
\end{itemize}

\subsubsection{Visualization of Optimized Gating Network}
We visualize several representative optimized gating networks after convergence with similar relative $L_2$ errors in Figures \ref{fig:BB_gate}, \ref{fig:BB4_gate_X} and \ref{fig:BB4_gate_M}, for the APINNs with two subnets, APINN4-X and APINN4-M, respectively. Note that the variance of this Boussinesq-Burger equation is smaller, so these models have similar performances.
The key observation is that the gate nets after optimization maintain the characteristics at initialization, especially for APINN-M.
Specifically, for APINN-M, the optimized gate networks do not change much from the initialization.
For APINN-X, although the position and slope of the interfaces between subdomains change, the optimized APINN-X is still partitioning the whole domain into four upper-to-bottom parts. 
Therefore, we have the following conclusions.
\begin{itemize}
\item Initialization is crucial to the success of APINN, which is reflected in the performance gaps between APINN-M and APINN-X, since the gate networks after optimization maintain the characteristics at initialization.
\item APINN with one kind of initialization can hardly be optimized into another kind. For instance, we seldom see gate nets of APINN-M are optimized to be similar to the decomposition of XPINNs.
\item These observations are consistent with our Theorem \ref{thm:2}, which states that a good initialization of the gate net contributes to better generalization, since the gate net does not need to change significantly from its initialization.
\item However, based on our extensive experiments, trainable gate nets still contribute to generalization, due to the positive fine-tuning effect, although it cannot optimize a MPINN-type APINN into a XPINN-type APINN and vice versa.
\end{itemize}
Furthermore, we visualize the optimization trajectory of the gating network for all subnets in the Boussinesq-Burger equation in Figure \ref{gif:BB} in the Appendix, where each snapshot is the gating net at epochs = 0, 10, 20, 30, 40, and 50. The change is fast and continuous.

\section{Summary}
In this paper, we propose the Augmented Physics-Informed Neural Networks (APINN) method, which employs a gate network for soft domain partition that can mimic the hard eXtended PINN (XPINN) domain decomposition and is trainable and fine-tunable. The gate network satisfying the partition-of-unity property averages several sub-networks as the output of APINN. Moreover, it adopts partial parameter sharing for sub-nets. It has the following advantages over the state-of-the-art generalized space-time domain decomposition based XPINN method:
\begin{itemize}
\item APINN does not include the complicated interface losses to maintain the continuity between different sub-networks (sub-PINNs) due to the gate network decomposing the entire domain in a soft way, which also contributes to better convergence and lower training loss.
\item The gate network can mimic the hard decomposition of XPINN, such that APINN enjoys the advantage of XPINN in that it can decompose the complicated target function into several simpler parts to reduce the complexity and improve the generalizability of each sub-network.
\item The trainable gate network enables fine-tuning the domain decomposition to discover a better function as well as domain decomposition for simpler parts, contributing to better generalization based on \cite{hu2021extended}.
\item The parameter sharing in APINN utilizes the essential idea that each sub-PINN is learning one part of the same target function, so that the commonality can be well captured by the shared part.
\item Each sub-networks in APINN takes advantage of all training samples within the domain to prevent over-fitting. By contrast, sub-networks in XPINN can only utilize part of the training samples.
\end{itemize}
All of the benefits are justified empirically on various PDEs and theoretically in \cite{hu2021extended} using the PINN generalization theory.
More specifically, we prove the generalization bound for APINNs with fixed and trainable get networks. Since APINNs with certain gate networks can recover PINN and XPINN, they have the advantages of the two models due to their trainability and flexibility. It is shown that APINN enjoys the benefit of general domain and function decomposition, which reduces the complexity of the optimized networks to improve generalization.
In terms of parallelization, APINN shares more data points as well as parameters than XPINNs, and thus can be more expensive than the XPINN method.

\section*{Acknowledgment}
A. D. Jagtap and G. E. Karniadakis would like to acknowledge the funding by  OSD/AFOSR  MURI  Grant  FA9550-20-1-0358, and the US Department of Energy (DOE) PhILMs  project  (DE-SC0019453).

\appendix
\section{Proof}
\subsection{Preliminary}
The proof depends on Rademacher complexity and covering number defined below.
\begin{definition}
(Rademacher Complexity). Let $S=\left\{x_i\right\}_{i=1}^n \subset \overline{\Omega}$ be a dataset containing $n$ samples. The Rademacher complexity of a function class $\mathcal{F}$ on $S$ is defined as
$\text{Rad}(\mathcal{F};S)=\mathbb{E}_{\epsilon}\left[\sup_{f\in\mathcal{F}} \frac{1}{n}\sum_{i=1}^n \epsilon_if(x_i)\right]$,
where $\epsilon_1,\dots,\epsilon_n$ are independent and identically distributed (i.i.d.)  random variables taking values uniformly in $\{-1,1\}$. 
\end{definition}
\begin{definition}
(Matrix Covering) We use $\mathcal{N}(U, \epsilon, \Vert \cdot \Vert)$ to denote the least cardinality of any subset $V \subset U$ that covers $U$ at scale $\epsilon$ with norm $\Vert \cdot \Vert$, i.e.,
$
\sup_{A \in U}\min_{B \in V} \Vert A - B \Vert \leq \epsilon.
$
\end{definition}
They are correlated as below.
\begin{lemma}
\label{lemma:A1}
\cite{bartlett2017spectrally}
Let $\mathcal{F}$ be a real-valued function class taking values in $[0,1]$, and assume that $\mathbf{0} \in \mathcal{F}$. Then
\begin{equation}
\text{Rad}\left(\mathcal{F};{ S}\right) \leq \inf _{\alpha>0}\left(\frac{4 \alpha}{\sqrt{n}}+\frac{12}{n} \int_{\alpha}^{\sqrt{n}} \sqrt{\log \mathcal{N}\left(\mathcal{F}_{ S}, \varepsilon,\|\cdot\|_{2,2}\right)} d \varepsilon\right),
\end{equation}
where $S$ is the dataset, and $\mathcal{F}_S$ is the set containing the image of the dataset $S$ under all mappings in $\mathcal{F}$.
\end{lemma}
We note that this lemma requires that the hypothesis is in the interval $[0,1]$. In practice, we can consider the class of truncated neural networks. More specifically, if the class of neural network is denoted $\mathcal{F}$, then we consider the following class:
\begin{equation}
\widehat{\mathcal{F}} = \mathcal{F}_+ + \mathcal{F}_{-},
\end{equation}
where $\mathcal{F}_+ = \left\{f: f \cap [0,1], f \in \mathcal{F}\right\}$ and $\mathcal{F}_- = \left\{f: f \cap [-1,0], f \in \mathcal{F}\right\}$. Then, the function class $\widehat{F}$ is bounded in the interval $[-1,1]$, which is suitable for prediction of the target function $u^*({\bx})$ which is bounded by 1. In addition, their Rademacher complexity have the relationship: $\text{Rad}(\widehat{\mathcal{F}}) \leq \text{Rad}(\mathcal{F}_+) + \text{Rad}(\mathcal{F}_-)$. Throughout this paper, we will adopt the truncated neural network function class unless specified.

\subsection{Proof of Theorem \ref{thm:1}}
\begin{proof}
We can abstract the proof into deriving the Rademacher complexity of the function class:
\begin{equation}
\mathcal{F}_G = \left\{\bx \mapsto \sum_{j=1}^m G(x)_j f_j(\bx) \quad \bigg|\quad f_j \in \mathcal{F}_j, G \ \text{fixed}\right\},
\end{equation}
where each $\mathcal{F}_j$ is a function class, e.g., that of multilayer networks corresponding to $E_j \circ h$. Using the property that $\text{Rad}(\mathcal{F}+\mathcal{G}) \leq \text{Rad}(\mathcal{F}) + \text{Rad}(\mathcal{G})$ and the fact that the multiplication of $G(\bx)_j$ is $\max_{x \in \partial \Omega}\Vert G(\bx) \Vert_\infty$-Lipschitz, we have
\begin{equation}
\begin{aligned}
\text{Rad}(\mathcal{F}_G) &\leq \sum_{j=1}^m\max_{\bx \in \partial \Omega}\Vert G(\bx)_j \Vert_{\infty}R_0(E_j \circ h).
\end{aligned}
\end{equation}
The proof is completed by the above inequality, the relationship between Rademacher complexity and generalization error, and Theorem 3.2 in \cite{hu2021extended}.
\end{proof}
\subsection{Proof of Theorem \ref{thm:2}}
\begin{proof}
Basically, we wish to derive the covering number $\mathcal{N}((\mathcal{F} \cdot \mathcal{G})_S, \epsilon)$ from the covering numbers $\mathcal{N}(\mathcal{F}_S, \epsilon)$ and $\mathcal{N}(\mathcal{G}_S, \epsilon)$, where $\mathcal{F}$ and $\mathcal{G}$ are function classes of neural networks, whose covering numbers have the forms
$\log\mathcal{N}(\mathcal{F}_S,\epsilon) = N(F)/\epsilon^2$ and 
$\log\mathcal{N}(\mathcal{G}_S,\epsilon) = N(G)/\epsilon^2$.

Take $\epsilon$ covers $\hat{\mathcal{F}}_S$ and $\hat{\mathcal{G}}_S$ for both $\mathcal{F}_S$ and $\mathcal{G}_S$, i.e., for all $f \in \mathcal{F}_S, g \in \mathcal{G}_S$, there exist $\hat{f} \in \hat{\mathcal{F}}_S, \hat{g} \in \hat{\mathcal{G}}_S$, such that $\Vert f - \hat{f} \Vert, \Vert g - \hat{g} \Vert \leq \epsilon$.
Using the inequality,
$
\Vert fg - \hat{f}\hat{g}\Vert \leq \Vert g\Vert\cdot\Vert f - \hat{f}\Vert + \Vert \hat{f}\Vert\cdot\Vert g - \hat{g}\Vert,
$ and our assumption on truncated neural network functions,
we know that
$
\mathcal{N}((\mathcal{F} \cdot \mathcal{G})_S, \epsilon) \leq  \mathcal{N}(\mathcal{F}_S, \epsilon) \cdot \mathcal{N}(\mathcal{G}_S, \epsilon),
$
since $\hat{\mathcal{F}}\cdot\hat{\mathcal{G}}$ is an $\epsilon$ cover of $(\mathcal{F}\cdot\mathcal{G})_S$.
Consequently,
$
\log \left(\mathcal{N}((\mathcal{F} \cdot \mathcal{G})_S, \epsilon)\right) \leq  \log\left(\mathcal{N}(\mathcal{F}_S, \epsilon)\right) \cdot \log\left( \mathcal{N}(\mathcal{G}_S, \epsilon)\right).$
By Lemma \ref{lemma:A1}, $\text{Rad}(\mathcal{F} \cdot \mathcal{G}) \leq O\left(\text{Rad}(\mathcal{F}) + \text{Rad}(\mathcal{G})\right)$. Combined with the relationship between Rademacher complexity and generalization error and Theorem 3.2 in \cite{hu2021extended}, we are done.
\end{proof}

\section{Optimization trajectory of the gating network for all subnets in the Boussinesq-Burger equation}
\begin{figure}[!h]
\centering
\includegraphics[scale=0.24]{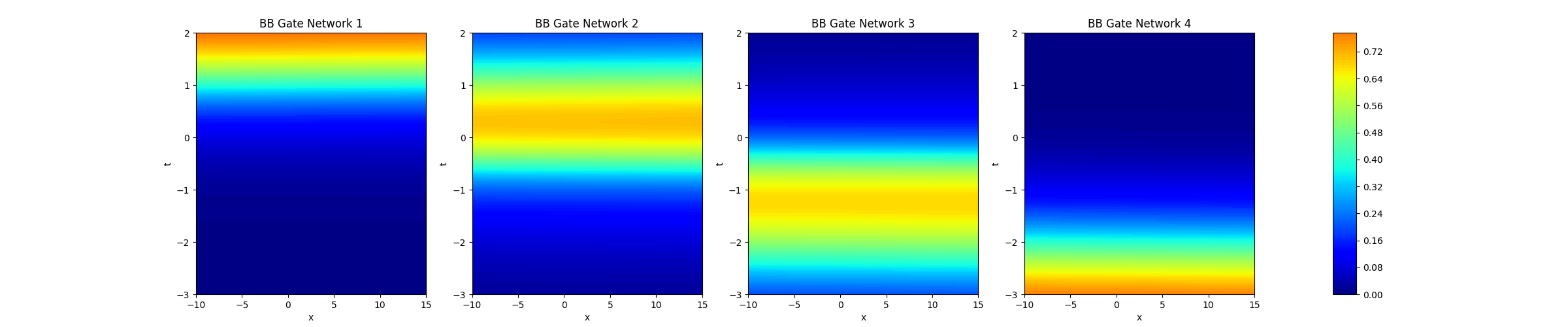}
\includegraphics[scale=0.24]{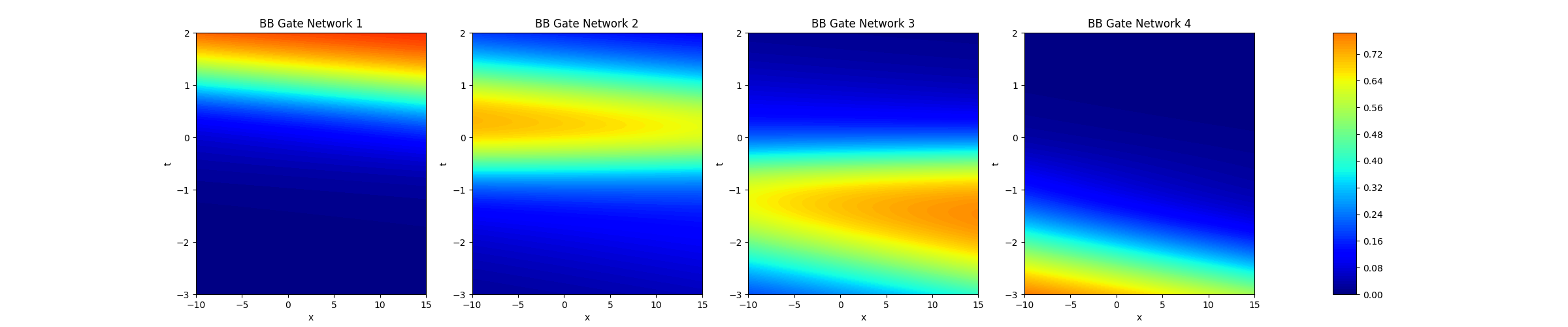}
\includegraphics[scale=0.24]{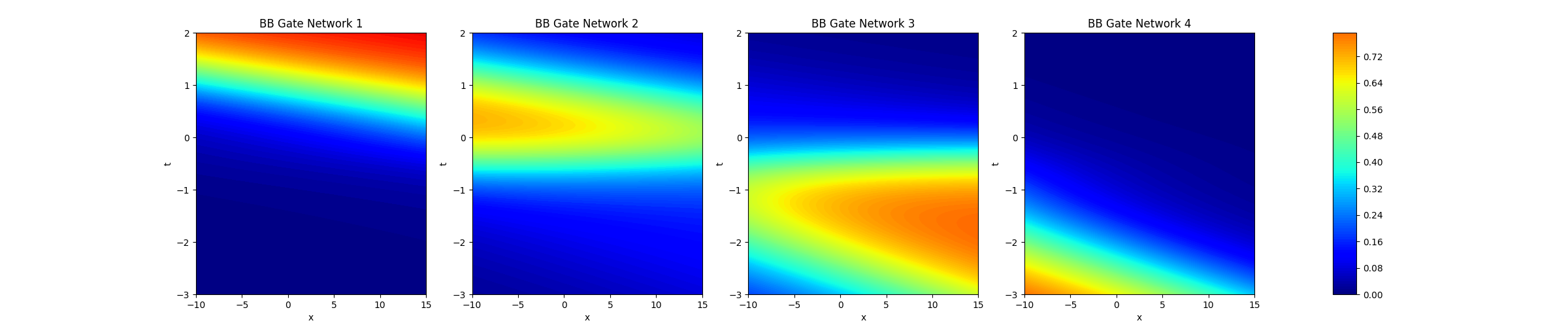}
\includegraphics[scale=0.24]{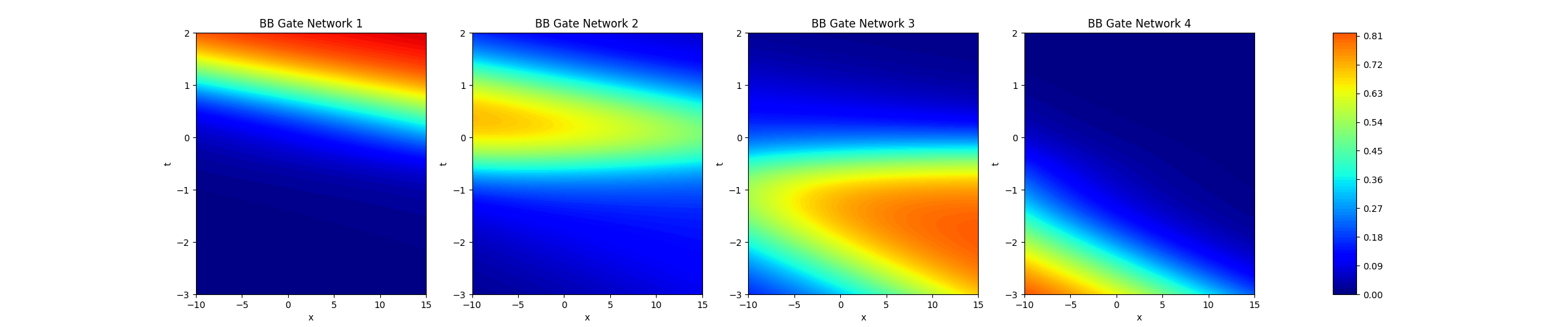}
\includegraphics[scale=0.24]{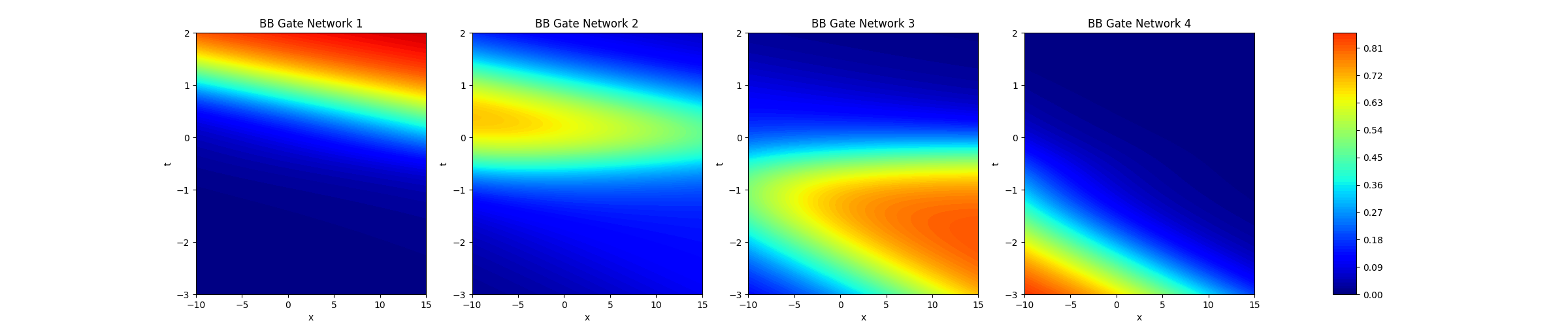}
\includegraphics[scale=0.24]{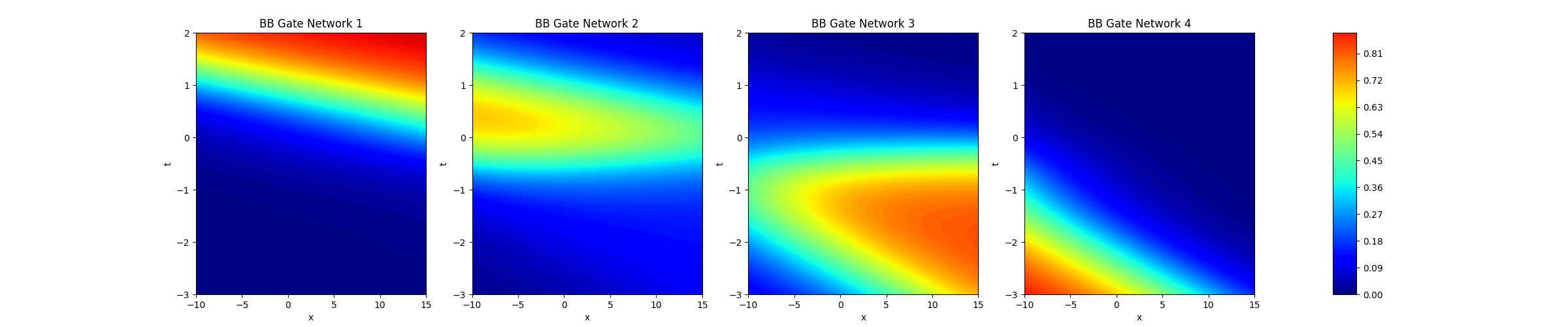}
\caption{The Boussinesq-Burger equation: (Top to bottom) Visualization of the gating network optimization trajectory via snapshots, at epoch = 0, 10, 20, 30, 40, 50.}
\label{gif:BB}
\end{figure}

The optimization trajectory of the gating network for all subnets in the Boussinesq-Burger equation is visualized in Figure \ref{gif:BB}, where each snapshots are the gating net at epoch = 0, 10, 20, 30, 40, and 50. 

\newpage
 \bibliographystyle{unsrtnat} 
 \bibliography{cas-refs}





\end{document}